\documentclass[lettersize,journal]{IEEEtran}

\usepackage{amsmath,amsfonts}
\usepackage{graphicx}
\usepackage{cite}
\usepackage{array}
\usepackage[caption=false,font=normalsize,labelfont=sf,textfont=sf]{subfig}
\usepackage{textcomp}
\usepackage{stfloats}
\usepackage{url}
\usepackage{verbatim}

\usepackage{booktabs}
\usepackage{multirow}
\usepackage{makecell}
\usepackage[table]{xcolor}
\usepackage{adjustbox}

\usepackage{float}
\usepackage{algorithm}
\usepackage{algpseudocode}

\floatstyle{ruled}
\restylefloat{algorithm}

\begin{document}

\title{DynFly: Dynamic-Aware Continuous Trajectory Generation for UAV Vision-Language Navigation in Urban Environments}


\author{Wen Jiang, Hanfang Liang, Li Wang, Kangyao Huang, Wang Xu,
Wei Fan, Jinyuan Liu, \\Shaoyu Liu, Hongwei Duan, Bin Xu,
Xiangyang Ji,~\IEEEmembership{Member,~IEEE},
and Huaping Liu,~\IEEEmembership{Fellow,~IEEE}

\thanks{Wen Jiang and Hanfang Liang contributed equally to this work and should be considered co-first authors.}

\thanks{
This work was supported by the National Natural Science Foundation of China under Grant No. 52502496, U22B2052 and the Natural Science Foundation of Chongqing, China under Grant No. CSTB2025NSCQ-GPX0413, and the National High Technology Research and Development Program of China under Grant No. 2020YFC1512501. (Corresponding authors: Bin Xu and Xiangyang Ji)}

\thanks{Wen Jiang, Li Wang, Wei Fan, Hongwei Duan, and Bin Xu are with the School of Mechanical Engineering, Beijing Institute of Technology, Beijing 100081, China. (e-mail: 3120235086@bit.edu.cn, wangli\_bit@bit.edu.cn,fanweixx@bit.edu.cn, 3220250437@bit.edu.cn, bitxubin@bit.edu.cn, }
\thanks{Kangyao Huang is with the Department of Computer Science and Technology, Tsinghua University, Beijing 100084, China.(e-mail: huangky22@mails.tsinghua.edu.cn)}

\thanks{Wang Xu is with Tsinghua University, Beijing 100084, China.(e-mail:xwjim812@126.com)}

\thanks{Li Wang is also with the Chongqing Innovation Center, Beijing Institute of Technology, Chongqing 401120, China.}

\thanks{Xiangyang Ji is with the Department of Automation, Tsinghua University, Beijing 100084, China.(e-mail: xyji@tsinghua.edu.cn)}
\thanks{Jinyuan Liu is with the School of Software, Dalian University of Technology, Dalian 116024, China.(e-mail: jinyuanliu@dlut.edu.cn)}
\thanks{Shaoyu Liu is with the School of Artificial Intelligence, Xidian University, Xi'an 710071, China.(e-mail: 23171110721@stu.xidian.edu.cn)}
\thanks{Hanfang Liang is Jianghan University, Wuhan 430056, China.(e-mail: lhf.liang@gmail.com)}
\thanks{Huaping Liu is with the Department of Computer Science and Technology, Tsinghua University, Beijing 100084, China (e-mail: hpliu@tsinghua.edu.cn).}

}

\markboth{Journal of \LaTeX\ Class Files,~Vol.~14, No.~8, August~2021}%
{Shell \MakeLowercase{\textit{et al.}}: A Sample Article Using IEEEtran.cls for IEEE Journals}


\maketitle
\begin{abstract}
Recent advances in multimodal large models have significantly improved UAV vision-language navigation (UAV-VLN) by enhancing high-level perception and reasoning. However, existing methods mainly focus on predicting discrete actions, local targets, or sparse waypoints, while the continuous transition from navigation intent to executable UAV motion remains weakly modeled. This motion-interface gap limits the continuity, stability, and executability of generated UAV trajectories. To address this gap, we propose DynFly, a dynamic-aware continuous trajectory generation framework that bridges high-level navigation reasoning and executable UAV motion. DynFly bridges high-level navigation intent and continuous UAV motion through a lightweight trajectory generation layer. Specifically, it represents expert trajectories in B-spline control-point space and employs a Spline-DiT generator to learn conditional trajectory generation via flow matching. Furthermore, we introduce UAV-oriented dynamic-aware supervision over position, finite-difference velocity, finite-difference acceleration, heading consistency, and local target alignment, enabling the generated trajectories to better satisfy UAV motion characteristics. And our trajectory generation framework can also be integrated with an existing UAV-VLN framework while preserving its original visual-language reasoning pipeline. Extensive experiments on the OpenUAV UAV-VLN benchmark show that DynFly improves both navigation performance and trajectory quality. On the Test Unseen Full split, DynFly improves the strongest baseline by 4.69 NDTW, 2.40 SDTW, 2.14\% SR and 4.87\% OSR, while reducing NE by 4.51 m.

\end{abstract}











\section{Introduction}
\label{sec:introduction}

Recent advances in multimodal large models have significantly improved UAV vision-language navigation (UAV-VLN)~\cite{DAI2026113330, GU2026112722, DEWANGAN2026113303} by enhancing visual perception, language understanding, and high-level navigation reasoning. Given multi-view visual observations and natural language instructions, modern UAV-VLN systems are capable of planning navigation goals in complex 3D environments~\cite{anderson2018vision,fried2018speakerFollower}. However, unlike ground robots~\cite{HE2024110511, MOHAMMADI2026112462}, UAVs operate in continuous three-dimensional space, where successful navigation requires not only selecting correct navigation targets but also generating executable flight trajectories that satisfy motion continuity and kinematic constraints. Therefore, transforming high-level navigation intent into continuous UAV motion has become an important problem in UAV-VLN.

Most existing UAV-VLN methods improve navigation performance by enhancing semantic understanding and spatial reasoning. Typical approaches introduce stronger visual encoders, geometry-enhanced scene representations, historical memory, or large vision-language models~\cite{liu2024eventgpteventstreamunderstanding} to improve target localization and navigation planning. Recent systems, including TravelUAV, AutoFly, and LongFly~\cite{wang2024towards,sun2026autofly,jiang2025longflylonghorizonuavvisionandlanguage}, further improve high-level reasoning ability through multimodal foundation models. Although these methods achieve encouraging navigation performance, their final outputs are still commonly represented as discrete actions, local targets, or sparse waypoint sequences. Such representations mainly answer where to fly, while the continuous transition between navigation targets remains weakly modeled. 

In UAV flight, this transition is critical because the vehicle cannot move by instantaneously jumping between sparse targets. Instead, it must follow a continuous motion process with coherent displacement, velocity, acceleration, and heading changes. 
Continuous trajectory generation therefore provides a more suitable motion representation for UAV navigation, as it explicitly models intermediate flight states and produces trajectories that are smoother and easier for downstream controllers to track.

\begin{figure*}[t]
    \centering
    \includegraphics[width=\linewidth]{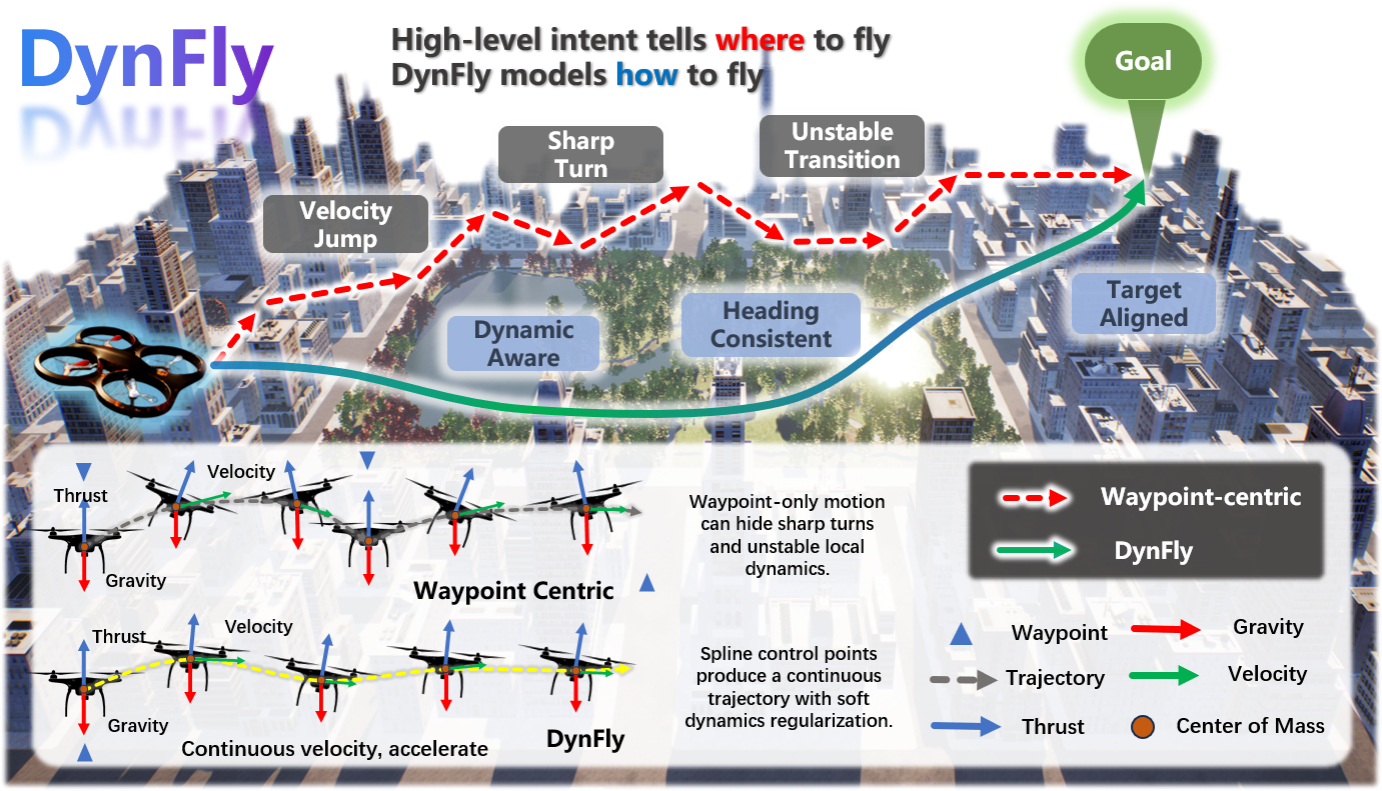}
    \caption{Motivation of the proposed dynamic-aware trajectory generation framework. Under the same visual-language navigation intent and goal, sparse waypoint prediction may produce under-constrained aerial motion with abrupt turns, velocity jumps, and unstable local transitions. DynFly converts the high-level intent into a continuous 3D UAV trajectory through dynamic-aware trajectory generation, yielding smoother motion, improved heading consistency, and better target alignment.}
    \label{fig:teaser}
\end{figure*}

This observation exposes a key limitation in current UAV-VLN systems, which lack an explicit motion interface between high-level navigation intent and continuous UAV flight execution. 

Motivated by this limitation, we propose DynFly, a dynamic-aware continuous trajectory generation framework for UAV vision-language navigation, as illustrated in Fig.~\ref{fig:teaser}. DynFly preserves existing visual-language reasoning modules and introduces an independent motion layer that converts high-level navigation intent into executable continuous UAV trajectories. Specifically, expert trajectories are represented in B-spline control-point space, where a lightweight Spline-DiT learns conditional trajectory generation through flow matching. The generated control points are then decoded into continuous trajectories and optimized using UAV-oriented dynamic-aware supervision over position, finite-difference velocity, finite-difference acceleration, heading consistency, and local target alignment. By introducing motion priors during trajectory generation, DynFly produces smoother, more stable, and physically executable UAV trajectories while preserving the original visual-language reasoning pipeline.

The main contributions are summarized as follows:
\begin{itemize}

    \item We propose DynFly, a dynamic-aware continuous trajectory generation framework that bridges high-level navigation reasoning and executable UAV motion. The proposed framework preserves the original visual-language reasoning pipeline while introducing an explicit trajectory generation layer for continuous UAV motion.
    \item We develop a spline-structured continuous trajectory generation framework that represents trajectories in B-spline control-point space and employs a lightweight Spline-DiT with flow matching to learn conditional trajectory generation, enabling the transformation of high-level navigation intent into continuous 3D UAV trajectories.
    \item We introduce UAV-oriented dynamic-aware supervision by incorporating motion priors over position, finite-difference velocity, finite-difference acceleration, heading consistency, and target alignment, leading to smoother, more stable, and physically executable UAV trajectories.
    \item Extensive experiments on UAV-VLN benchmarks demonstrate that DynFly consistently improves both navigation performance and trajectory quality, while further validating the effectiveness and compatibility of the proposed trajectory generation framework.
    
\end{itemize}

\section{Related Work}
\label{sec:related_work}

\subsection{Vision-Language Navigation for UAVs}

VLN aims to guide embodied agents to navigate according to natural-language instructions and visual observations. Early works such as R2R and Speaker-Follower established the basic VLN setting by connecting instruction understanding, visual state encoding, and sequential action prediction~\cite{fried2018speakerFollower,liu2024navagentmultiscaleurbanstreet}. Later methods, including HAMT, and DUET, improved navigation performance through cross-modal representation learning, semantic grounding, history modeling, and topological reasoning~\cite{chen2022hamt,chen2022duet}. With the development of large language models and vision-language models, VLN has further evolved toward embodied reasoning. MapGPT, MapNav, and AgentVLN improve long-horizon navigation through language reasoning, semantic maps, and structured planning~\cite{chen2024mapgpt,zhang2025mapnav,xin2026agentvln}, while vision-language-action models such as PaLM-E, RT-2, and OpenVLA extend visual-language understanding to robot action generation~\cite{driess2023palme,zitkovich2023rt2,kim2024openvla,jing2026tempovlalearningspeedcontrollablevisionlanguageaction}. These advances have also promoted UAV vision-language navigation. Representative systems, including AerialVLN~\cite{liu2023aerialvln}, TravelUAV~\cite{wang2024towards}, OpenFly~\cite{gao2025openfly}, FlightGPT~\cite{cai2025flightgpt}, LongFly~\cite{jiang2025longflylonghorizonuavvisionandlanguage}, and AutoFly~\cite{sun2026autofly}, improve UAV navigation through stronger perception, spatial reasoning, historical memory, or multimodal foundation models. Although these methods substantially improve high-level navigation capability, most of them still represent navigation outputs as discrete actions, local targets, or sparse waypoint sequences. Such representations mainly determine where the UAV should fly, while the conversion from navigation intent to executable continuous UAV motion remains weakly modeled.

\subsection{Continuous UAV Trajectory Modeling}

Continuous trajectory modeling plays an important role in UAV flight because UAV motion is closely related to velocity, acceleration, heading changes, and trajectory smoothness. Traditional UAV trajectory planning methods commonly employ B-splines, minimum-snap optimization, or kinodynamic planning to generate smooth and dynamically feasible trajectories~\cite{babaei2018uavBspline,zhou2020egoplanner,burke2021fastSplineMinimumSnap}. Recent label-free UAV trajectory prediction work has also explored pseudo-label supervision and dynamics-aware motion modeling; for example, Liang et al.~\cite{liang2025labelfreelonghorizon3duav} generate pseudo trajectory labels and incorporate UAV motion dynamics to predict long-horizon 3D trajectories without explicit trajectory annotations. Representative systems such as EGO-Planner and PE-Planner further demonstrate that spline-based trajectory representations are effective for real-time UAV planning and local replanning~\cite{zhou2020egoplanner,qiu2024peplanner}. However, these methods mainly rely on geometric maps, local sensing, or optimization objectives and are not designed to model visual-language navigation intent. More recently, generative action models such as Diffusion Policy, $\pi_0$, and FlowMP have shown that diffusion models and flow matching can generate continuous action sequences from visual or task conditions~\cite{chi2023diffusionpolicy,black2024pi0,nguyen2025flowmp}. Although these methods demonstrate the potential of generative modeling for continuous control, they generally operate in action spaces and do not explicitly exploit the continuous curve structure of UAV trajectories. Overall, existing research mainly follows two separate directions: UAV-VLN methods focus on high-level navigation reasoning and answer where to fly, while trajectory planning and generative action models focus on continuous motion generation and answer how to move. The motion interface between these two levels remains underexplored. DynFly addresses this gap by introducing a dynamic-aware continuous trajectory generation framework that explicitly models the transition from navigation intent to executable UAV motion.

\section{Method}
\label{sec:method}

\subsection{Problem Formulation}
This paper studies the motion-interface problem in UAV-VLN, where high-level visual-language intent is converted into a short-horizon continuous UAV trajectory. 

At time $t$, $\mathcal{R}_t=\{I_t^{\mathrm{front}}, I_t^{\mathrm{rear}}, I_t^{\mathrm{left}}, I_t^{\mathrm{right}}, I_t^{\mathrm{down}}\}$ denotes the multi-view RGB observations, a language instruction $\mathcal{L}=\{w_i\}_{i=1}^{N}$, and the pose of the UAV $\mathbf{S}_t=[x_t,y_t,z_t,\varphi_t,\theta_t,\psi_t]$.

The visual-language front-end produces a context representation and a local navigation intent:
\begin{equation}
    (\mathbf{h}_t,\mathbf{g}_t)
    =
    \Phi_{\mathrm{vl}}(\mathcal{R}_t,\mathcal{L},\mathbf{S}_t),
\end{equation}
where $\mathbf{h}_t$ denotes the visual-language context and 
$\mathbf{g}_t\in\mathbb{R}^{3}$ is the local 3D target consumed by the motion layer.

Given $(\mathbf{h}_t,\mathbf{g}_t)$, DynFly predicts a future 3D trajectory over the next $T$ steps:
\begin{equation}
    \hat{\mathcal{Y}}_t =
    \{\hat{\mathbf{p}}_{t+1},\ldots,\hat{\mathbf{p}}_{t+T}\},
    \quad
    \hat{\mathbf{p}}_{t+i}\in\mathbb{R}^{3},
\end{equation}
with the supervised trajectory denoted as 
$\mathcal{Y}^{*}_t=\{\mathbf{p}^{*}_{t+1},\ldots,\mathbf{p}^{*}_{t+T}\}$.
The local motion generation function is formulated as
\begin{equation}
    \hat{\mathcal{Y}}_t =
    \Psi_{\mathrm{motion}}(\mathbf{h}_t,\mathbf{g}_t).
\end{equation}
Unlike waypoint-centric prediction, DynFly models short-horizon UAV motion in B-spline control-point space and supervises the decoded trajectory with motion-aware trajectory losses.

\subsection{System Overview}

\begin{figure*}[t]
    \centering
    \includegraphics[width=\linewidth]{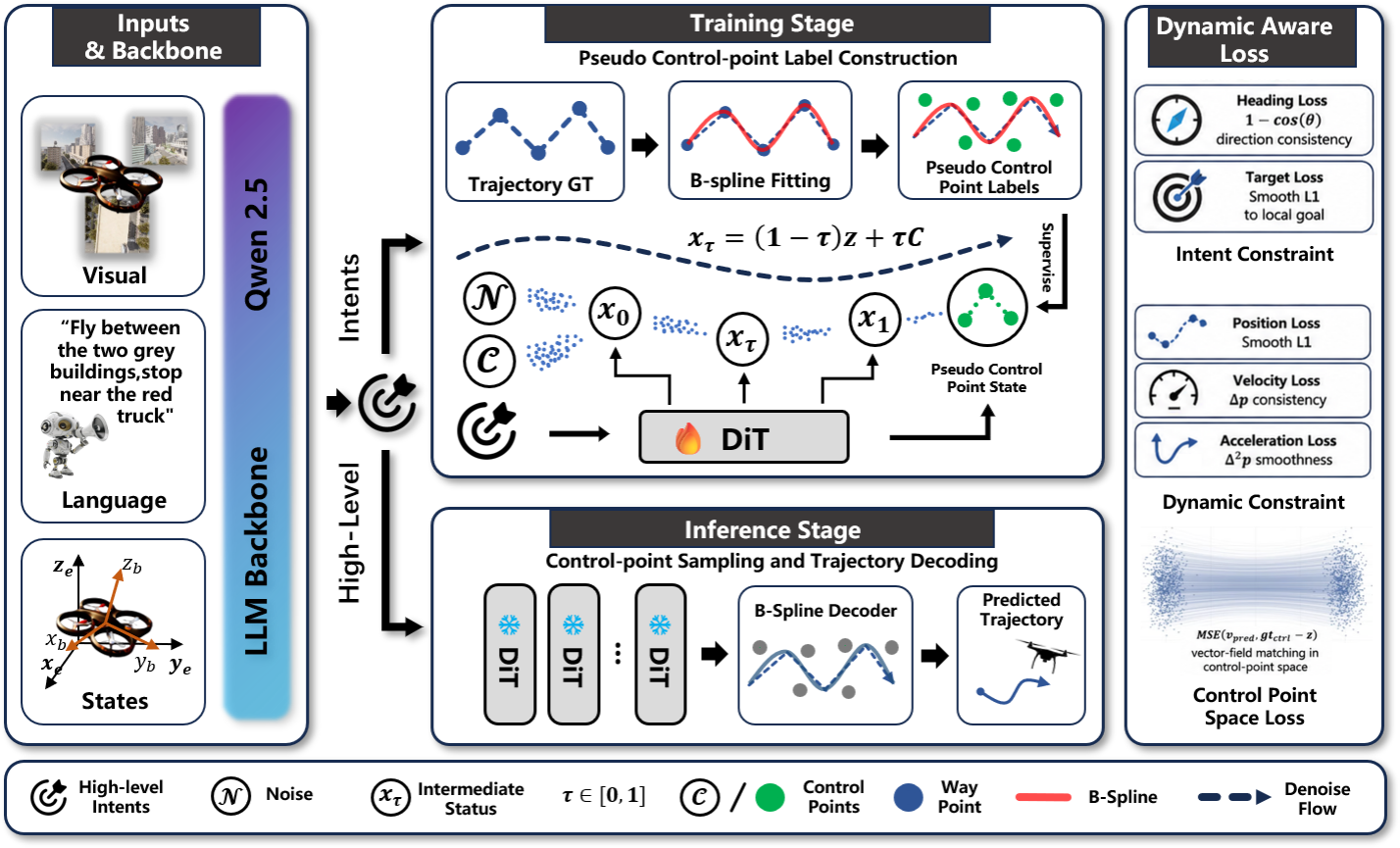}
    \caption{
    Overall architecture and workflow of DynFly. The Qwen2.5-3B visual-language front-end produces a context representation and a local navigation intent from multi-view observations, language instruction, and UAV state. The proposed motion layer then generates B-spline control points through conditional flow matching and decodes them into a continuous UAV trajectory with trajectory-level dynamic-aware supervision.
    }
    \label{fig:framework}
\end{figure*}

Fig.~\ref{fig:framework} shows the overall architecture and workflow of DynFly. 
The central idea is to separate high-level visual-language reasoning from low-level continuous motion generation, while connecting them through a compact motion interface. 
Given multi-view RGB observations $\mathcal{R}_t$, language instruction $\mathcal{L}$, and UAV state $\mathbf{S}_t$, the visual-language front-end produces a context representation and a local navigation intent:
\begin{equation}
    (\mathbf{h}_t,\mathbf{g}_t)
    =
    \mathcal{F}_{\Theta}(\mathcal{R}_t,\mathcal{L},\mathbf{S}_t),
\end{equation}
where $\mathbf{h}_t$ denotes the visual-language context and $\mathbf{g}_t\in\mathbb{R}^{3}$ is the local 3D target consumed by the trajectory generator. 
The front-end is responsible for high-level semantic and spatial reasoning, namely where the UAV should fly next, while DynFly converts this intent into how to fly through a continuous short-horizon trajectory.

In our implementation, the visual-language front-end is built on Qwen2.5-3B. 
Multi-view RGB observations are encoded by an EVA ViT-G vision tower with a CLIP image processor, and the selected high-level visual features are projected into the multimodal hidden space through a projector. 
This front-end provides visual-language context and local navigation intent for the subsequent motion layer. 
DynFly therefore focuses on the motion-interface problem: transforming high-level intent into spline-structured continuous UAV trajectories.

To address the motion-interface gap in UAV-VLN, DynFly keeps the visual-language front-end responsible for semantic grounding and local intent prediction, while shifting the main modeling effort to the subsequent motion-generation module. The proposed module then transforms the front-end intent into spline-structured continuous UAV trajectories.

During training, expert waypoint trajectories are converted into B-spline pseudo control-point labels and used to supervise a Spline-DiT generator in control-point space through flow matching. 
During inference, the same conditioned generator samples B-spline control points from noise, and a B-spline decoder maps the generated control points into a continuous short-horizon UAV trajectory. 
This layered design directly targets the motion-generation interface that is often hidden in waypoint-centric UAV-VLN systems.

The trajectory condition encoder projects the visual-language context, local target, and flow-matching time into a shared hidden space:
\begin{equation}
\begin{aligned}
\mathbf{c}_t^{\mathrm{vis}}
&=
\mathrm{Proj}_{\mathrm{vis}}(\mathbf{h}_t),\quad
\mathbf{c}_t^{\mathrm{tar}}
=
\mathrm{MLP}_{\mathrm{tar}}(\mathbf{g}_t),\\
\mathbf{c}^{\tau}
&=
\mathrm{MLP}_{\tau}\!\left(\mathrm{Embed}(\tau)\right).
\end{aligned}
\label{eq:condition}
\end{equation}
where $\tau\in[0,1]$ denotes the flow-matching generation time. 
These conditions are fused into the global condition vector for Spline-DiT:
\begin{equation}
    \mathbf{h}_{\tau}
    =
    \mathrm{Fuse}
    \left(
    \mathbf{c}_t^{\mathrm{vis}}
    +
    \mathbf{c}_t^{\mathrm{tar}}
    +
    \mathbf{c}^{\tau}
    \right).
\end{equation}

Conditioned on $\mathbf{h}_{\tau}$, the Spline-DiT predicts a vector field in B-spline control-point space rather than directly predicting independent waypoints. 
The generated control points are decoded into a 3D trajectory using a precomputed B-spline basis matrix. 
The decoded trajectory is supervised with position matching, first-order displacement consistency, second-order motion variation, direction consistency, and local target alignment. 
Overall, DynFly serves as a trajectory generation layer that bridges high-level navigation reasoning and executable UAV motion.

\subsection{Pseudo Control-Point Label Generation}

Given the current front-end context and local navigation target, DynFly generates a 3D trajectory that respects local UAV motion continuity. Instead of directly predicting discrete waypoints, we reformulate trajectory generation as conditional flow-field modeling in B-spline control-point space. The first step is to convert discrete trajectory labels into continuous-curve control points, which serve as pseudo control-point labels.

Waypoint-centric methods usually represent the expert trajectory as
\begin{equation}
    \mathcal{Y}^{*}
    =
    [\mathbf{y}^{*}_{0},\mathbf{y}^{*}_{1},\ldots,
    \mathbf{y}^{*}_{T-1}],
    \quad
    \mathbf{y}^{*}_{t}\in\mathbb{R}^{3}.
\end{equation}
This representation is intuitive, but discrete waypoints do not explicitly describe the continuous motion between adjacent points. When neighboring waypoints are far apart or the trajectory contains sharp turns, a model may learn position matching while ignoring velocity variation, acceleration smoothness, and direction continuity. For UAVs, this can produce trajectories with discontinuities, velocity jumps, or overly aggressive local turns.

We therefore convert the expert trajectory into B-spline control-point space. B-spline curves provide local continuity and smoothness, and their control points compactly describe the geometry of an entire trajectory. The expert trajectory is mapped to a set of control points
\begin{equation}
    \mathbf{C}^{*}
    =
    [\mathbf{c}^{*}_{0},\mathbf{c}^{*}_{1},\ldots,
    \mathbf{c}^{*}_{N_{\mathrm{cp}}-1}]
    \in\mathbb{R}^{N_{\mathrm{cp}}\times 3},
\end{equation}
where $N_{\mathrm{cp}}$ is the number of control points. We call $\mathbf{C}^{*}$ a pseudo control-point label because it is fitted from expert waypoints rather than directly annotated as a separate sensor signal. The label is no longer an isolated waypoint sequence, but a dynamic-aware training target obtained through continuous curve parameterization.

\begin{figure*}[t]
    \centering
    \includegraphics[width=0.95\linewidth]{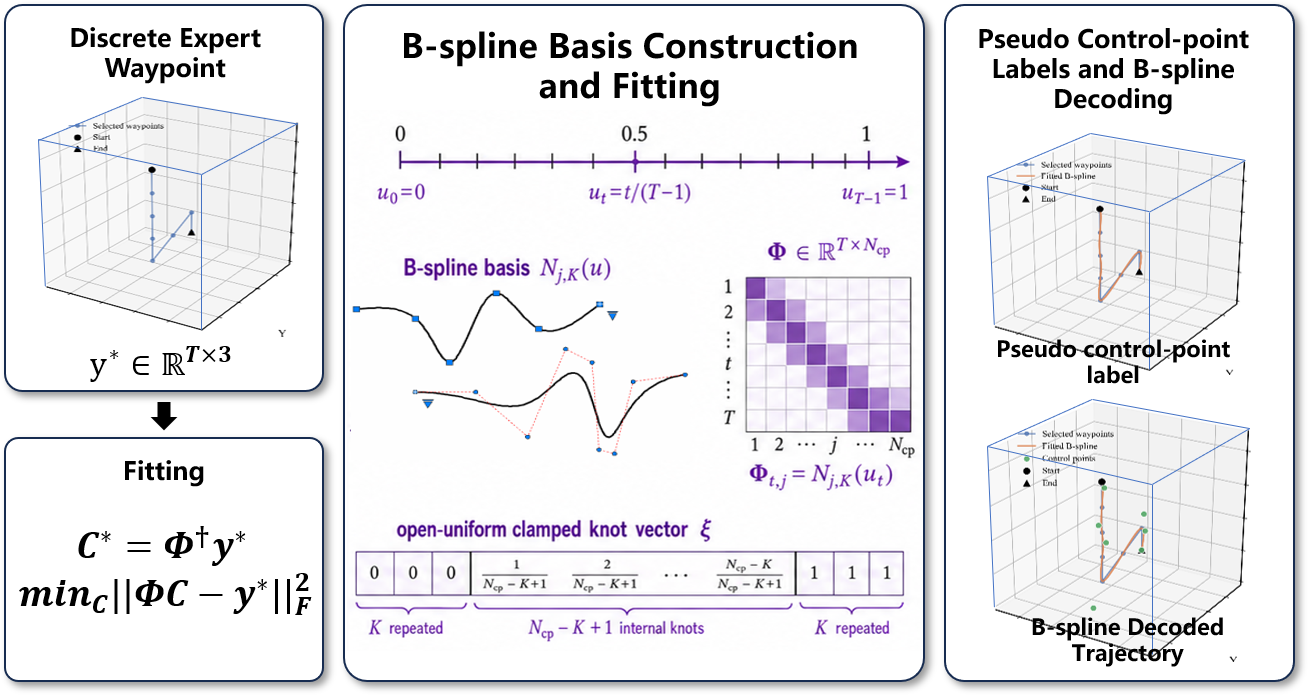}
    \caption{
    Pseudo control-point label generation. Discrete expert waypoints are fitted with an open-uniform clamped B-spline basis to obtain pseudo control-point labels 
    $\mathbf{C}^{*}=\Phi^{\dagger}\mathcal{Y}^{*}$. These labels provide a compact control-point-space target for flow matching and can be decoded back into continuous B-spline trajectories.
    }
    \label{fig:dynamic_label_generation}
\end{figure*}

Fig.~\ref{fig:dynamic_label_generation} summarizes this conversion without relying on a formula-heavy view. Raw expert waypoints are first aligned to a normalized trajectory time grid, a fixed B-spline basis is constructed, and the control polygon is fitted so that the decoded curve reconstructs the expert trajectory. The resulting control points become compact pseudo control-point labels for the subsequent flow-generation module.

Let the B-spline degree be $K$, the number of control points be $N_{\mathrm{cp}}$, and the number of discrete trajectory steps be $T$. We construct an open-uniform clamped knot vector
\begin{equation}
    \boldsymbol{\xi}
    =
    [\xi_0,\xi_1,\ldots,\xi_{N_{\mathrm{cp}}+K}],
\end{equation}
with length $N_{\mathrm{cp}}+K+1$. In a clamped B-spline, the first and last knots are repeated $K+1$ times, which stabilizes the curve near the endpoints. For normalized trajectory time $u\in[0,1]$, the B-spline basis functions $N_{j,K}(u)$ are computed by the standard Cox-de Boor recursion. Evaluating the basis at
\begin{equation}
    u_t = \frac{t}{T-1},
    \quad t=0,1,\ldots,T-1,
\end{equation}
gives a basis matrix
\begin{equation}
    \mathbf{\Phi}\in\mathbb{R}^{T\times N_{\mathrm{cp}}},
    \quad
    \Phi_{t,j}=N_{j,K}(u_t).
\end{equation}
Rows of $\mathbf{\Phi}$ are normalized so that $\sum_{j=0}^{N_{\mathrm{cp}}-1}\Phi_{t,j}=1$. Any control-point sequence $\mathbf{C}\in\mathbb{R}^{N_{\mathrm{cp}}\times3}$ can then be decoded into a trajectory:
\begin{equation}
    \hat{\mathcal{Y}}
    =
    \mathbf{\Phi}\mathbf{C},
    \quad
    \hat{\mathbf{y}}_t
    =
    \sum_{j=0}^{N_{\mathrm{cp}}-1}\Phi_{t,j}\mathbf{c}_j.
\end{equation}
Because B-spline basis functions have local support and smooth recursive structure, the decoded trajectory naturally introduces structural coupling between neighboring trajectory points.

Training data provides an expert waypoint trajectory $\mathcal{Y}^{*}$, while the generator learns in control-point space. We therefore solve for control-point labels whose decoded curve reconstructs the expert trajectory:
\begin{equation}
    \mathbf{C}^{*}
    =
    \arg\min_{\mathbf{C}}
    \left\|
    \mathbf{\Phi}\mathbf{C}
    -
    \mathcal{Y}^{*}
    \right\|_{F}^{2}.
\end{equation}
The closed-form least-squares solution is obtained by the Moore-Penrose pseudo-inverse:
\begin{equation}
    \mathbf{C}^{*}
    =
    \mathbf{\Phi}^{\dagger}\mathcal{Y}^{*},
    \quad
    \mathbf{\Phi}^{\dagger}\in\mathbb{R}^{N_{\mathrm{cp}}\times T}.
\end{equation}
When $N_{\mathrm{cp}}>T$, as in the default $N_{\mathrm{cp}}=8,T=8,K=3$ configuration, this fitting problem is underdetermined. The pseudo-inverse therefore selects the deterministic minimum-norm control-point label among the least-squares solutions; it should be interpreted as a compact training target construction rather than a unique physical trajectory parameterization.
This converts waypoint labels into control-point labels. Compared with direct waypoint supervision, the resulting pseudo control-point label compresses the whole trajectory shape, encourages continuous curves after decoding, and provides a more suitable generation space for flow matching.

\subsection{DiT-Based Control Point Flow Generation}

After obtaining the pseudo control-point label $\mathbf{C}^{*}$, DynFly models trajectory generation as conditional control-point generation. Given the front-end context $\mathbf{h}_t$ and local target $\mathbf{g}_t$, the model generates control points
\begin{equation}
    \hat{\mathbf{C}}\in\mathbb{R}^{N_{\mathrm{cp}}\times3},
    \quad
    \hat{\mathcal{Y}}=\mathbf{\Phi}\hat{\mathbf{C}}.
\end{equation}
The core distribution is
\begin{equation}
    p_{\theta}(\mathbf{C}\mid \mathbf{h}_t,\mathbf{g}_t).
\end{equation}
We use flow matching to learn a conditional vector field in control-point space:
\begin{equation}
    \mathbf{v}_{\theta}(\mathbf{X}_{\tau},\mathbf{h}_{\tau}),
\end{equation}
where $\mathbf{X}_{\tau}\in\mathbb{R}^{N_{\mathrm{cp}}\times3}$ is the intermediate control-point state at flow time $\tau\in[0,1]$, and $\mathbf{h}_{\tau}\in\mathbb{R}^{d}$ is a condition vector produced from the front-end context, target, and time features.

The trajectory generator reuses the visual-language context produced by the front-end. If the front-end exposes visual tokens or a feature map, these features are pooled into a compact context vector; if it exposes a single context embedding, that embedding is used directly. The context vector is projected to the DiT hidden dimension:
\begin{equation}
    \bar{\mathbf{f}}_{\mathrm{img}}
    =
    \frac{1}{M}\sum_{m=1}^{M}\mathbf{f}_m,
    \quad
    \mathbf{e}_{\mathrm{img}}
    =
    \phi_{\mathrm{img}}(\bar{\mathbf{f}}_{\mathrm{img}}).
\end{equation}
This keeps the upstream visual-language model compact: the trajectory generator uses the front-end context as a global semantic condition rather than introducing an additional task-specific visual decoder head.

The local target $\mathbf{g}\in\mathbb{R}^{3}$ is encoded by a lightweight MLP:
\begin{equation}
    \mathbf{e}_{\mathrm{tar}}
    =
    \phi_{\mathrm{tar}}(\mathbf{g}),
    \quad
    \phi_{\mathrm{tar}}:\mathbb{R}^{3}\rightarrow\mathbb{R}^{d}.
\end{equation}
Flow matching also requires the model to predict vector fields at arbitrary $\tau$. We use sinusoidal timestep embedding followed by an MLP:
\begin{equation}
    \mathbf{e}_{\tau}
    =
    \phi_{\tau}(\gamma(\tau)).
\end{equation}
The image, target, and time conditions are fused by addition followed by a small fusion network:
\begin{equation}
    \mathbf{h}_{\tau}
    =
    \phi_{\mathrm{fuse}}
    (
    \mathbf{e}_{\mathrm{img}}
    +
    \mathbf{e}_{\mathrm{tar}}
    +
    \mathbf{e}_{\tau}
    ).
\end{equation}

\begin{figure}[t]
    \centering
    \includegraphics[width=0.5\linewidth]{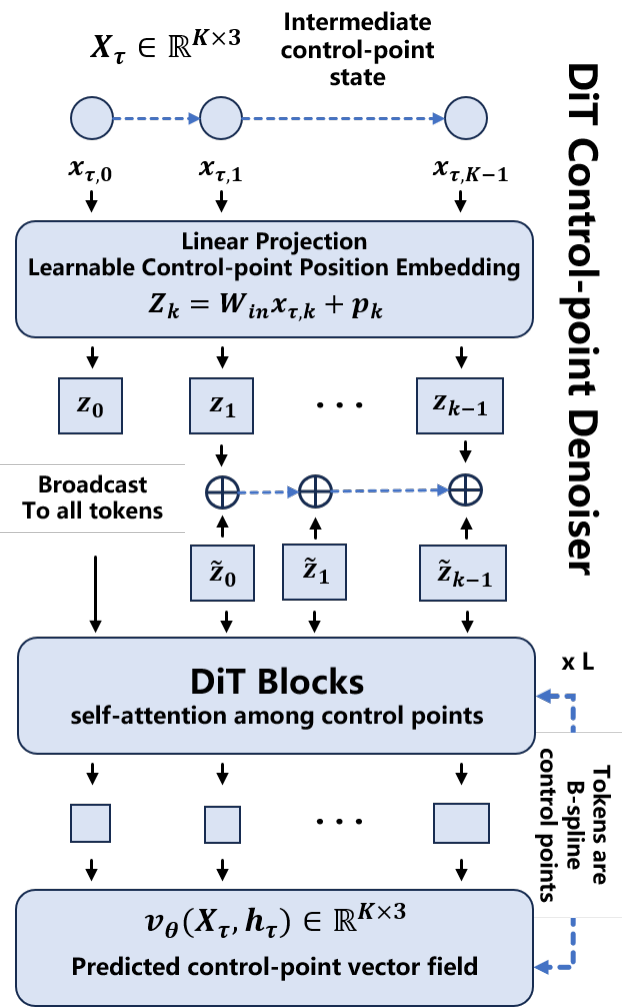}
    \caption{DiT-based control-point flow generation. The condition encoder fuses visual context, the local target, and flow time into a global condition $\mathbf{h}_{\tau}$. The intermediate control-point state $\mathbf{X}_{\tau}$ is embedded as B-spline control-point tokens with learnable positional encodings, processed by DiT blocks, and mapped to the vector field $\mathbf{v}_{\theta}(\mathbf{X}_{\tau},\mathbf{h}_{\tau})$. Flow-matching training supervises the field from Gaussian noise to pseudo control-point labels, while inference performs Euler sampling in control-point space and decodes the final control points with the B-spline basis.}
    \label{fig:spline_dit}
\end{figure}

At flow time $\tau$, the input control-point state is projected into hidden tokens and augmented with learnable control-point positional embeddings:
\begin{equation}
    \mathbf{z}_{j}
    =
    \mathbf{W}_{\mathrm{in}}\mathbf{x}_{\tau,j}
    +
    \mathbf{p}_{j},
    \quad j=0,1,\ldots,N_{\mathrm{cp}}-1.
\end{equation}
The global condition is normalized and broadcast to every control-point token. The token sequence is processed by a Transformer encoder, and a final LayerNorm plus linear layer predicts the control-point vector field:
\begin{equation}
    \mathbf{v}_{\theta}
    (
    \mathbf{X}_{\tau},
    \mathbf{h}_{\tau}
    )
    \in
    \mathbb{R}^{N_{\mathrm{cp}}\times3}.
\end{equation}
This module can be viewed as a lightweight DiT-style denoiser, but its tokens are B-spline control points rather than image patches, and it predicts a flow vector field rather than image noise. Fig.~\ref{fig:spline_dit} summarizes the full control-point flow generation process, including condition encoding, DiT-based control-point denoising, flow-matching supervision, and inference-time Euler sampling followed by B-spline decoding.

To learn a continuous transformation from noise to trajectory control points, we sample Gaussian noise
\begin{equation}
    \mathbf{Z}\sim\mathcal{N}(\mathbf{0},\mathbf{I}),
    \quad
    \mathbf{Z}\in\mathbb{R}^{N_{\mathrm{cp}}\times3},
\end{equation}
and a random flow time $\tau\sim\mathcal{U}(0,1)$. Given pseudo control-point label $\mathbf{C}^{*}$, the linear interpolation path is
\begin{equation}
    \mathbf{X}_{\tau}
    =
    (1-\tau)\mathbf{Z}
    +
    \tau\mathbf{C}^{*}.
\end{equation}
Equivalently, the learned field approximates the conditional flow
\begin{equation}
    \frac{d\mathbf{X}_{\tau}}{d\tau}
    =
    \mathbf{v}_{\theta}(\mathbf{X}_{\tau},\mathbf{h}_{\tau}),
    \quad
    \mathbf{X}_{0}=\mathbf{Z},\quad
    \mathbf{X}_{1}=\mathbf{C}^{*}.
\end{equation}
The target vector field is
\begin{equation}
    \mathbf{u}_{\tau}
    =
    \mathbf{C}^{*}-\mathbf{Z}.
\end{equation}
The flow-matching objective is
\begin{equation}
    \mathcal{L}_{\mathrm{FM}}
    =
    \mathbb{E}
    \left[
    \left\|
    \mathbf{v}_{\theta}(\mathbf{X}_{\tau},\mathbf{h}_{\tau})
    -
    (\mathbf{C}^{*}-\mathbf{Z})
    \right\|_2^2
    \right].
\end{equation}
For trajectory-level supervision during training, we form the implemented auxiliary clean control-point estimate from the same noise anchor:
\begin{equation}
    \hat{\mathbf{C}}
    =
    \mathbf{Z}
    +
    \mathbf{v}_{\theta}(\mathbf{X}_{\tau},\mathbf{h}_{\tau}),
    \quad
    \hat{\mathcal{Y}}
    =
    \mathbf{\Phi}\hat{\mathbf{C}}.
\end{equation}
This estimate follows the implemented auxiliary decoder path for trajectory-level losses. At inference, the model does not access $\mathbf{C}^{*}$. It initializes $\mathbf{C}^{(0)}\sim\mathcal{N}(\mathbf{0},\mathbf{I})$, sets $\tau_i=i/S$ for $i=0,\ldots,S-1$, constructs $\mathbf{h}_{\tau_i}$ from the same front-end context, local target, and the current time embedding, and applies Euler integration:
\begin{equation}
    \mathbf{C}^{(i+1)}
    =
    \mathbf{C}^{(i)}
    +
    \Delta\tau\,
    \mathbf{v}_{\theta}
    (
    \mathbf{C}^{(i)},
    \mathbf{h}_{\tau_i}
    ),
    \quad
    \Delta\tau=\frac{1}{S}.
\end{equation}
After $S$ steps, $\mathbf{C}^{(S)}$ is decoded by $\mathbf{\Phi}$ to obtain the predicted UAV trajectory.

\subsection{Loss Design}

\begin{figure}[t]
    \centering
    \includegraphics[width=0.8\linewidth]{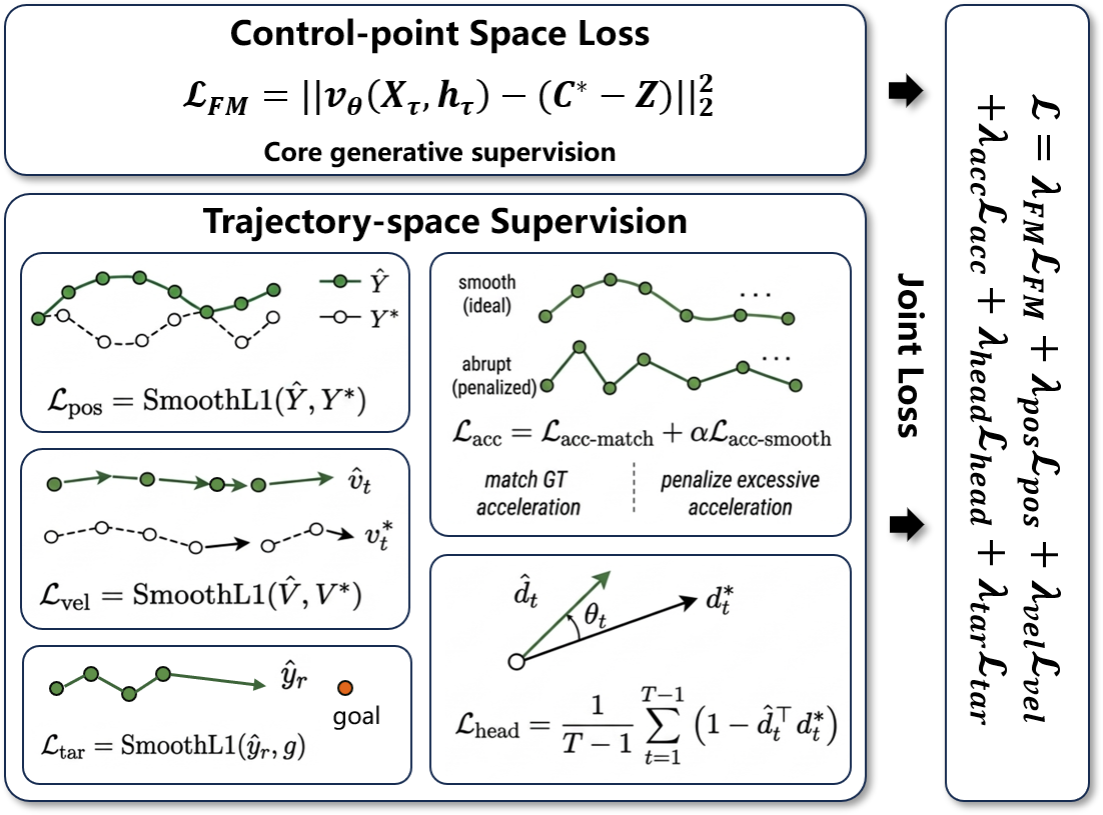}
    \caption{Joint loss design for Spline-DiT trajectory generation. Flow matching provides control-point-space supervision from noisy control points to pseudo control-point labels. The predicted control points are decoded by the B-spline basis into a continuous trajectory, where position, finite-difference velocity, finite-difference acceleration, heading, and target-anchoring losses form the dynamic-aware trajectory supervision and improve goal responsiveness.}
    \label{fig:loss_design}
\end{figure}

To make the model learn both control-point generation and trajectory-level dynamic consistency, DynFly uses a joint objective consisting of flow matching, position, velocity, acceleration, heading, and target losses. 
Fig.~\ref{fig:loss_design} illustrates how the control-point-space supervision and trajectory-space supervision are connected through B-spline decoding.

For each sampled training tuple, the per-sample flow-matching loss is $\mathcal{L}_{\mathrm{FM}}$. Although the generator learns in control-point space, the final task is trajectory prediction. We therefore decode $\hat{\mathbf{C}}$ into $\hat{\mathcal{Y}}$ and apply a position loss:
\begin{equation}
    \mathcal{L}_{\mathrm{pos}}
    =
    \mathrm{SmoothL1}
    (
    \hat{\mathcal{Y}},
    \mathcal{Y}^{*}
    ).
\end{equation}
All trajectory points are aligned by the local prediction horizon used by the B-spline decoder. The velocity and acceleration terms are finite-difference constraints over this decoded trajectory. These finite-difference terms capture local displacement and curvature patterns that are physically relevant to UAV motion, and can therefore regularize real trajectory shapes in addition to matching waypoint positions. To match local motion trends, we approximate velocity with first differences:
\begin{equation}
    \hat{\mathbf{v}}_t
    =
    \hat{\mathbf{y}}_{t+1}
    -
    \hat{\mathbf{y}}_{t},
    \quad
    \mathbf{v}^{*}_t
    =
    \mathbf{y}^{*}_{t+1}
    -
    \mathbf{y}^{*}_{t},
\end{equation}

And define
\begin{equation}
    \mathcal{L}_{\mathrm{vel}}
    =
    \mathrm{SmoothL1}
    (
    \hat{\mathbf{V}},
    \mathbf{V}^{*}
    ).
\end{equation}
Acceleration is approximated by second differences:
\begin{equation}
    \hat{\mathbf{a}}_t
    =
    \hat{\mathbf{y}}_{t+2}
    -
    2\hat{\mathbf{y}}_{t+1}
    +
    \hat{\mathbf{y}}_{t},
    \quad
    \mathbf{a}^{*}_t
    =
    \mathbf{y}^{*}_{t+2}
    -
    2\mathbf{y}^{*}_{t+1}
    +
    \mathbf{y}^{*}_{t}.
\end{equation}

The acceleration loss matches expert acceleration and suppresses overly large predicted second-order changes:
\begin{equation}
    \mathcal{L}_{\mathrm{acc}}
    =
    \mathrm{SmoothL1}
    (
    \hat{\mathbf{A}},
    \mathbf{A}^{*}
    )
    +
    \alpha
    \frac{1}{T-2}
    \sum_{t=0}^{T-3}
    \|\hat{\mathbf{a}}_t\|_2^2.
\end{equation}

The implementation uses $\alpha=0.1$ for this acceleration penalty.
To further avoid overly aggressive local motion, we impose upper-bound constraints on the predicted finite-difference velocity and acceleration.

Heading consistency is computed from normalized velocity directions:
\begin{equation}
    \hat{\mathbf{d}}_t
    =
    \frac{\hat{\mathbf{v}}_t}{\|\hat{\mathbf{v}}_t\|_2+\epsilon},
    \quad
    \mathbf{d}^{*}_t
    =
    \frac{\mathbf{v}^{*}_t}{\|\mathbf{v}^{*}_t\|_2+\epsilon},
\end{equation}
\begin{equation}
    \mathcal{L}_{\mathrm{head}}
    =
    1
    -
    \frac{1}{T-1}
    \sum_{t=0}^{T-2}
    \hat{\mathbf{d}}_t^{\top}\mathbf{d}^{*}_t 
\end{equation}
where $\epsilon$ is a small constant for numerical stability.

This loss directly constrains the local flight direction and helps avoid trajectories that are positionally close but directionally inconsistent.

The local target $\mathbf{g}$ is anchored at an intermediate trajectory index
\begin{equation}
    r=\min(4,T-1),
\end{equation}
with target loss
\begin{equation}
    \mathcal{L}_{\mathrm{tar}}
    =
    \mathrm{SmoothL1}
    (
    \hat{\mathbf{y}}_r,
    \mathbf{g}
    ).
\end{equation}
The index follows the implemented target anchor in the trajectory layer and places the local target near the early-middle part of the short horizon. This does not force the final point to equal the local target. Instead, it encourages the short-horizon trajectory to move toward the target while leaving later points available for smoothing, local adjustment, or motion variation.

The final training objective is
\begin{equation}
\begin{aligned}
    \mathcal{L} ={}&
    \lambda_{\mathrm{FM}}\mathcal{L}_{\mathrm{FM}}
    +
    \lambda_{\mathrm{pos}}\mathcal{L}_{\mathrm{pos}}
    +
    \lambda_{\mathrm{vel}}\mathcal{L}_{\mathrm{vel}}
    +
    \lambda_{\mathrm{acc}}\mathcal{L}_{\mathrm{acc}} \\
    &+
    \lambda_{\mathrm{head}}\mathcal{L}_{\mathrm{head}}
    +
    \lambda_{\mathrm{tar}}\mathcal{L}_{\mathrm{tar}} .
\end{aligned}
\end{equation}
In the reported training setting, the weights are $\lambda_{\mathrm{FM}}=1.0$, $\lambda_{\mathrm{pos}}=1.0$, $\lambda_{\mathrm{vel}}=0.2$, $\lambda_{\mathrm{acc}}=0.05$, $\lambda_{\mathrm{head}}=0.05$, and $\lambda_{\mathrm{tar}}=0.1$. These dynamic-aware terms guide the decoded trajectory toward smoother local displacement, acceleration consistency, heading consistency, and target responsiveness.

\section{Experiments}
\label{sec:experiments}

\subsection{Experimental Setup}

\begin{table*}[t]
\caption{Results on the Test Seen Set across Full/Easy/Hard splits. The low-data block uses 25\% of training data and the full-data block uses 100\% of training data. The Human row is included as a seen-set reference and is not an algorithmic baseline.}
\label{tab:seen_results}
\centering

\footnotesize

\setlength{\tabcolsep}{3.2pt}

\renewcommand{\arraystretch}{1.08}

\begin{tabular*}{\textwidth}{@{\extracolsep{\fill}}lcccccccccccc}

\toprule
\multirow{2}{*}{Method} &
\multicolumn{4}{c}{\textbf{Full}} &
\multicolumn{4}{c}{\textbf{Easy}} &
\multicolumn{4}{c}{\textbf{Hard}} \\
\cmidrule(lr){2-5} \cmidrule(lr){6-9} \cmidrule(lr){10-13}
& NE$\downarrow$ & SR$\uparrow$ & OSR$\uparrow$ & SPL$\uparrow$
& NE$\downarrow$ & SR$\uparrow$ & OSR$\uparrow$ & SPL$\uparrow$
& NE$\downarrow$ & SR$\uparrow$ & OSR$\uparrow$ & SPL$\uparrow$ \\
\midrule
Human & \textbf{14.15} & \textbf{94.51} & \textbf{94.51} & \textbf{77.84}
& \textbf{11.68} & \textbf{95.44} & \textbf{95.44} & \textbf{76.19}
& \textbf{17.16} & \textbf{93.37} & \textbf{93.37} & \textbf{79.85} \\
\midrule
\multicolumn{13}{l}{\textbf{25\% Training Data}} \\
TravelUAV~\cite{wang2024towards}
& 132.59 & 11.59 & 24.50 & 10.45
& 85.75 & 14.11 & 30.35 & 12.34
& 181.19 & 8.98 & 18.43 & 8.48 \\
OpenVLN~\cite{lin2025openvlnopenworldaerialvisionlanguage}
& 125.97 & 14.39 & 28.03 & 12.94
& 87.96 & 15.22 & 30.64 & 13.31
& 175.54 & 13.32 & 24.62 & 12.55 \\
SpatialFly~\cite{jiang2026spatialflygeometryguidedrepresentationalignment} 
& 113.31 & 16.84 & 38.36 & 14.58
& 70.13 & 18.60 & 48.15 & 16.12
& 162.90 & 14.82 & 27.02 & 12.81 \\
\textbf{DynFly (Ours)}
& \textbf{63.96} & \textbf{32.93} & \textbf{64.25} & \textbf{27.24}
& \textbf{44.63} & \textbf{34.70} & \textbf{68.07} & \textbf{27.19}
& \textbf{86.15} & \textbf{30.91} & \textbf{59.85} & \textbf{27.31} \\
\midrule
\multicolumn{13}{l}{\textbf{100\% Training Data}} \\
Random Action
& 222.20 & 0.14 & 0.21 & 0.07
& 142.07 & 0.26 & 0.39 & 0.13
& 320.12 & 0.00 & 0.00 & 0.00 \\
Fixed Action
& 188.61 & 2.27 & 8.16 & 1.40
& 121.36 & 3.48 & 11.48 & 2.14
& 270.69 & 0.79 & 4.09 & 0.49 \\
CMA
& 135.73 & 8.37 & 18.72 & 7.90
& 84.89 & 11.48 & 24.52 & 10.68
& 197.77 & 4.57 & 11.65 & 4.51 \\
TravelUAV~\cite{wang2024towards}
& 106.28 & 16.10 & 44.26 & 14.30
& 68.78 & 18.84 & 47.61 & 16.39
& 152.04 & 12.76 & 40.16 & 11.76 \\
TravelUAV-DA~\cite{wang2024towards}
& 98.66 & 17.45 & 48.87 & 15.76
& 66.40 & 20.26 & 51.23 & 18.10
& 138.04 & 14.02 & 45.98 & 12.90 \\
NavFoM~\cite{zhang2025embodiednavigationfoundationmodel}
& 93.05 & 29.17 & 49.24 & 25.03
& 58.98 & 32.91 & 53.16 & 27.87
& 143.83 & 23.58 & 43.40 & 20.80 \\
LongFly~\cite{jiang2025longflylonghorizonuavvisionandlanguage}
& 60.02 & 36.39 & 65.87 & 31.07
& 38.10 & 38.52 & 71.90 & 31.24
& 85.20 & 33.94 & 58.94 & 30.88 \\
SpatialFly~\cite{jiang2026spatialflygeometryguidedrepresentationalignment} 
& 58.05 & 38.54 & 68.62 & \textbf{33.43}
& 34.41 & 39.59 & 74.14 & \textbf{33.61}
& 84.76 & 37.33 & 62.27 & 33.22 \\
\textbf{DynFly (Ours)}
& \textbf{53.66} & \textbf{40.62} & \textbf{69.89} & 33.33
& \textbf{33.46} & \textbf{42.74} & \textbf{74.27} & 33.35
& \textbf{76.85} & \textbf{38.18} & \textbf{64.85} & \textbf{33.30} \\
\bottomrule

\end{tabular*}

\end{table*}

\subsubsection{Datasets and Experimental Environment}

We evaluate DynFly on the OpenUAV UAV-VLN benchmark. The dataset consists of Train, Test Seen, Test Unseen Map, and Test Unseen Object splits, where the test sets are further divided into Easy and Hard subsets. The training set contains 9,152 trajectories, while the Test Seen, Test Unseen Map, and Test Unseen Object sets contain 1,418, 958, and 629 trajectories, respectively. All experiments are conducted in the AirSim simulator, which provides physics-based UAV simulation and realistic visual environments~\cite{airsim2017fsr}.

\subsubsection{Evaluation Metrics}

We report the standard navigation metrics, including normalized dynamic time warping (NDTW), success weighted by dynamic time warping (SDTW), navigation error (NE), success rate (SR), oracle success rate (OSR), and success weighted by path length (SPL). The DTW-based metrics are computed as
\begin{equation}
\begin{aligned}
\mathrm{NDTW}(P,G)
&=
\exp\!\left(
-\frac{\mathrm{DTW}(P,G)}
{\eta |G|}
\right),\\
\mathrm{SDTW}
&=
\mathbf{1}_{\mathrm{succ}}
\cdot
\mathrm{NDTW}(P,G).
\end{aligned}
\label{eq:ndtw}
\end{equation}
where $\eta=3.0$ and $\mathbf{1}_{\mathrm{succ}}$ indicates whether the navigation is successful.

In addition to the standard navigation metrics, we evaluate trajectory quality using SmoothMean, SmoothVar, TurnMean, and TurnVar. Smoothness is measured using the second-order difference of adjacent trajectory points,
\begin{equation}
    \mathbf{a}_i
    =
    \hat{\mathbf{p}}_{i+2}
    -
    2\hat{\mathbf{p}}_{i+1}
    +
    \hat{\mathbf{p}}_{i},
\end{equation}
where SmoothMean and SmoothVar denote the mean and variance of $\|\mathbf{a}_i\|_2$, respectively. Turning stability is measured by the angle between two consecutive motion directions,
\begin{equation}
    \theta_i
    =
    \arccos
    \left(
    \frac{
    (\hat{\mathbf{p}}_{i+1}-\hat{\mathbf{p}}_i)^{\top}
    (\hat{\mathbf{p}}_{i+2}-\hat{\mathbf{p}}_{i+1})
    }{
    (\|\hat{\mathbf{p}}_{i+1}-\hat{\mathbf{p}}_i\|_2+\epsilon)
    (\|\hat{\mathbf{p}}_{i+2}-\hat{\mathbf{p}}_{i+1}\|_2+\epsilon)
    }
    \right),
\end{equation}
where TurnMean and TurnVar denote the mean and variance of $\theta_i$, respectively. In implementation, the cosine value is clipped to [-1,1] before applying $\arccos$.

\subsubsection{Implementation Details}

The framework uses Qwen2.5-3B as the visual-language front-end and a frozen EVA ViT-G as the vision encoder. DynFly predicts a seven-step 3D trajectory represented by eight cubic B-spline control points and uses a Spline-DiT with a hidden size of 512, six Transformer blocks, and eight attention heads for control-point generation. The model is trained on eight NVIDIA RTX 4090 GPUs with a batch size of 8, a learning rate of $5\times10^{-4}$, and six training epochs using bf16 mixed precision. The dynamic-aware loss weights are set to $\lambda_{\mathrm{vel}}=0.2$, $\lambda_{\mathrm{acc}}=0.05$, $\lambda_{\mathrm{head}}=0.05$, and $\lambda_{\mathrm{tar}}=0.1$. During inference, trajectories are generated using 10 Euler integration steps.

\subsection{ Comparison With State-of-the-Art Methods} 

We compare DynFly with representative UAV-VLN methods, including Random Action, Fixed Action, CMA, TravelUAV and TravelUAV-DA~\cite{wang2024towards}, OpenVLN~\cite{lin2025openvlnopenworldaerialvisionlanguage}, NavFoM~\cite{zhang2025embodiednavigationfoundationmodel}, LongFly~\cite{jiang2025longflylonghorizonuavvisionandlanguage}, and SpatialFly~\cite{jiang2026spatialflygeometryguidedrepresentationalignment}. For a fair comparison, LongFly, SpatialFly, and DynFly use the same Qwen2.5-3B visual-language front-end and are trained and evaluated on the same dataset under the same training and evaluation settings. This enables a fair evaluation of the performance improvements brought by DynFly.

\subsubsection{ Quantitative Evaluation on the OpenUAV-Seen Dataset}

As shown in Table~\ref{tab:seen_results}, we further evaluate DynFly on the OpenUAV Test Seen dataset under two training settings, including 25\% Low-Data and 100\% Full-Data. Table~\ref{tab:seen_results} reports the complete quantitative results on the Full, Easy, and Hard splits, and provides an overall comparison of representative methods on the Seen dataset.

Under the 25\% Low-Data setting, DynFly achieves the best NE, SR, OSR, and SPL on the Full, Easy, and Hard splits. Compared with SpatialFly, DynFly reduces NE from 113.31\,m to 63.96\,m on the Full split, while improving SR from 16.84\% to 32.93\%, OSR from 38.36\% to 64.25\%, and SPL from 14.58\% to 27.24\%. Similar improvements are also observed on the Easy and Hard splits. For example, on the Hard split, DynFly reduces NE from 162.90\,m to 86.15\,m while improving SR to 30.91\%. These results indicate that DynFly is able to learn a stable continuous trajectory generation process even with limited training data, leading to better navigation performance.

Under the 100\% Full-Data setting, DynFly continues to achieve consistent improvements. On the Full split, DynFly improves SR from 38.54\% to 40.62\% while reducing NE from 58.05\,m to 53.66\,m compared with SpatialFly. On the Hard split, NE is further reduced from 84.76\,m to 76.85\,m, while SR and OSR increase to 38.18\% and 64.85\%, respectively. Although SPL remains comparable to SpatialFly, DynFly consistently achieves better navigation success and lower navigation error. These results suggest that the proposed dynamic-aware motion interface further improves navigation performance on top of a strong visual-language navigation framework.

Overall, DynFly consistently improves navigation performance under both limited-data and full-data settings.  The improvements are reflected not only in a single metric but also in navigation success, navigation accuracy, and path efficiency. These results further validate the effectiveness of the proposed dynamic-aware motion interface for UAV vision-language navigation.

\subsubsection{ Quantitative Evaluation on the OpenUAV-Unseen Dataset}

\begin{table*}[t]
\caption{Results on the Test Unseen Set across Full/Easy/Hard splits. NDTW, SDTW, SR, OSR, and SPL are higher-is-better; NE is lower-is-better. Dashes indicate unavailable metrics rather than zero scores.}
\label{tab:unseen_results}
\centering
\scriptsize
\setlength{\tabcolsep}{2.5pt}
\begin{adjustbox}{max width=\textwidth}
\begin{tabular}{lcccccccccccccccccc}
\toprule
\multirow{2}{*}{Method} &
\multicolumn{6}{c}{\textbf{Full}} &
\multicolumn{6}{c}{\textbf{Easy}} &
\multicolumn{6}{c}{\textbf{Hard}} \\
\cmidrule(lr){2-7} \cmidrule(lr){8-13} \cmidrule(lr){14-19}
& NDTW$\uparrow$ & SDTW$\uparrow$ & NE$\downarrow$ & SR$\uparrow$ & OSR$\uparrow$ & SPL$\uparrow$
& NDTW$\uparrow$ & SDTW$\uparrow$ & NE$\downarrow$ & SR$\uparrow$ & OSR$\uparrow$ & SPL$\uparrow$
& NDTW$\uparrow$ & SDTW$\uparrow$ & NE$\downarrow$ & SR$\uparrow$ & OSR$\uparrow$ & SPL$\uparrow$ \\
\midrule
Random Action
& -- & -- & 225.64 & 0.06 & 0.06 & 0.06
& -- & -- & 164.66 & 0.19 & 0.19 & 0.19
& -- & -- & 280.58 & 0.00 & 0.00 & 0.00 \\
Fixed Action
& -- & -- & 193.30 & 1.76 & 5.36 & 1.09
& -- & -- & 140.33 & 3.19 & 8.08 & 1.88
& -- & -- & 245.96 & 0.85 & 3.08 & 0.55 \\
CMA
& -- & -- & 147.27 & 4.98 & 12.41 & 4.74
& -- & -- & 102.54 & 8.03 & 17.52 & 7.52
& -- & -- & 191.30 & 2.76 & 7.53 & 2.71 \\
TravelUAV~\cite{wang2024towards}
& -- & -- & 130.60 & 11.41 & 31.13 & 10.45
& -- & -- & 96.27 & 12.47 & 33.31 & 11.29
& -- & -- & 167.49 & 10.62 & 28.91 & 9.80 \\
NavFoM~\cite{zhang2025embodiednavigationfoundationmodel}
& -- & -- & 118.34 & 15.63 & 30.46 & 14.21
& -- & -- & 89.77 & 16.98 & 32.22 & 15.35
& -- & -- & 155.69 & 14.35 & 27.79 & 13.16 \\
LongFly~\cite{jiang2025longflylonghorizonuavvisionandlanguage}
& 9.31 & 5.57 & 91.84 & 24.19 & 43.86 & 20.84
& 9.09 & 5.31 & 69.16 & 22.89 & 43.24 & 18.66
& 9.51 & 5.81 & 112.02 & 25.36 & 44.41 & 22.76 \\
SpatialFly~\cite{jiang2026spatialflygeometryguidedrepresentationalignment} 
& 11.20 & 6.50 & 87.82 & 25.46 & 44.41 & 21.76
& 10.20 & 6.04 & \textbf{64.49} & 25.17 & 44.18 & 20.36
& 12.08 & 6.91 & 108.56 & 25.71 & 44.63 & 23.01 \\
\textbf{DynFly (Ours)}
& \textbf{15.89} & \textbf{8.90} & \textbf{83.31} & \textbf{27.60} & \textbf{49.28} & \textbf{23.57}
& \textbf{14.21} & \textbf{7.96} & 64.85 & \textbf{26.10} & \textbf{46.72} & \textbf{21.26}
& \textbf{17.38} & \textbf{9.73} & \textbf{99.73} & \textbf{28.93} & \textbf{51.55} & \textbf{25.63} \\
\bottomrule
\end{tabular}
\end{adjustbox}
\end{table*}

As shown in Table~\ref{tab:unseen_results}, we first evaluate DynFly on the OpenUAV Test Unseen dataset. The dataset contains the Full, Easy, and Hard splits, which are used to evaluate navigation performance in unseen environments. Table~\ref{tab:unseen_results} reports the complete quantitative results, and provides an overall comparison of the six evaluation metrics for representative methods.

On the Full split, DynFly achieves the best performance on all six metrics, including NDTW, SDTW, NE, SR, OSR, and SPL. Compared with the strongest baseline, SpatialFly, DynFly improves NDTW from 11.20 to 15.89 and SDTW from 6.50 to 8.90, while reducing NE from 87.82\,m to 83.31\,m. At the same time, SR, OSR, and SPL increase to 27.60\%, 49.28\%, and 23.57\%, respectively. These results indicate that the proposed dynamic-aware motion interface improves trajectory matching, navigation success, and path efficiency within the same UAV-VLN framework.

The improvement is more evident on the Hard split. Compared with SpatialFly, DynFly reduces NE from 108.56\,m to 99.73\,m and achieves the best results on NDTW, SDTW, SR, OSR, and SPL. This suggests that in more challenging unseen environments, improving high-level visual-language reasoning alone is not sufficient. Explicitly modeling the transition from high-level navigation intent to continuous UAV motion provides more stable performance gains.

On the Easy split, DynFly still achieves the best results on NDTW, SDTW, SR, OSR, and SPL, while NE is slightly higher than that of SpatialFly. This indicates that existing spatial representation methods are already able to achieve low navigation error in relatively simple scenarios, whereas the main advantage of DynFly lies in improving trajectory matching, navigation success, and path efficiency. Overall, the results on the Test Unseen dataset demonstrate the effectiveness of DynFly in unseen environments and show that the proposed dynamic-aware motion interface improves the conversion from high-level navigation intent to executable continuous UAV trajectories.

\subsubsection{ Performance on Unseen Map and Unseen Object
}
\begin{figure*}[t]
    \centering
    \includegraphics[width=\linewidth]{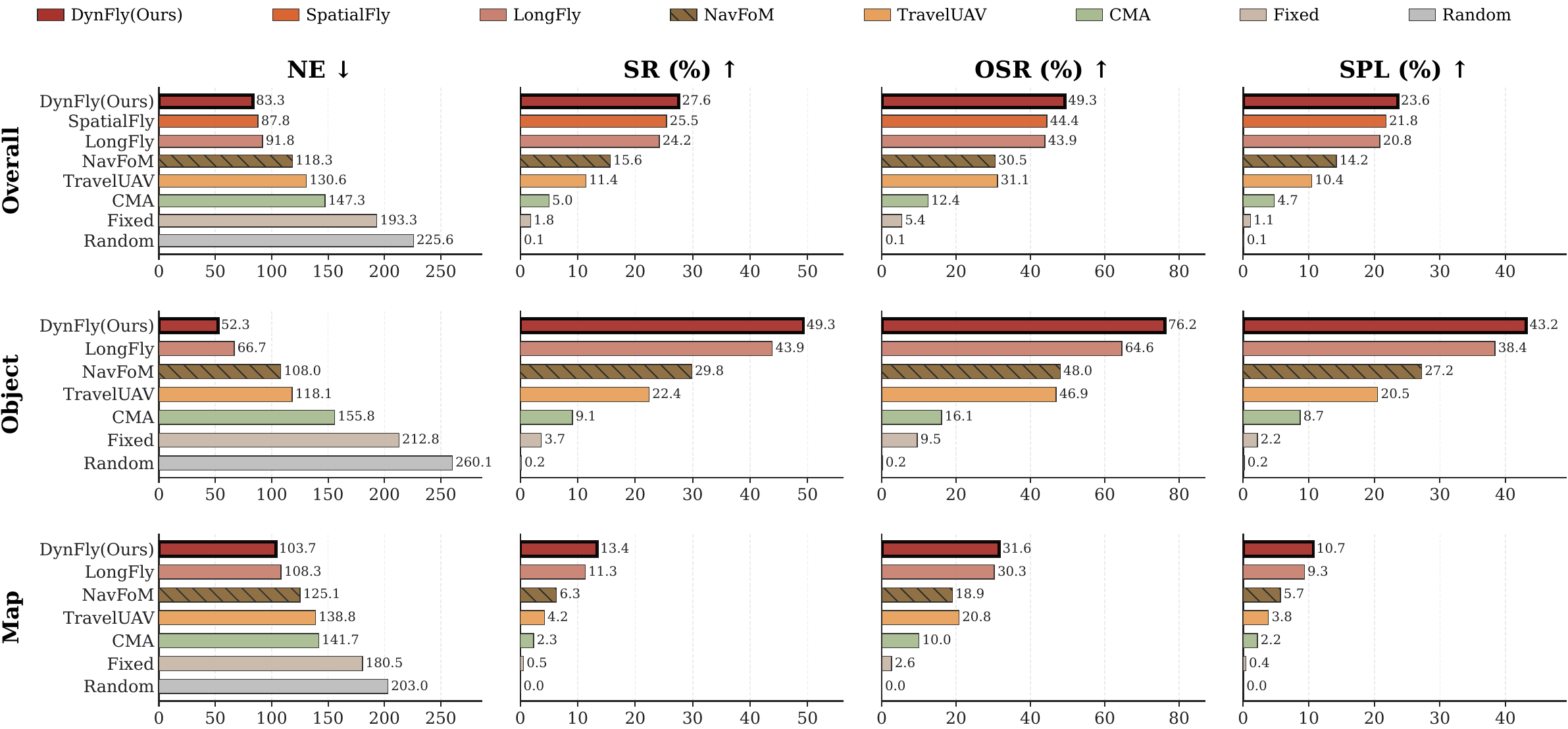}
    \caption{Full-split comparison across Unseen Overall, Unseen Object, and Unseen Map settings. DynFly improves most navigation metrics across the three unseen protocols, with especially strong gains on unseen-object generalization.}
    \label{fig:bar_full}
\end{figure*}

Figure~\ref{fig:bar_full} provides a visual comparison of the Full split under the Overall, Unseen Object, and Unseen Map evaluation protocols. Overall, DynFly achieves the best overall performance across the three settings, obtaining the lowest NE together with the highest SR, OSR, and SPL. These results show that the proposed dynamic-aware continuous trajectory generation method consistently improves the overall navigation performance of UAV vision-language navigation.

A comparison between the two unseen scenarios further shows that DynFly achieves larger improvements on the Unseen Object split, with better navigation success, path efficiency, and lower navigation error. This indicates that the proposed method can generate more stable continuous flight trajectories when object categories change, leading to better object-level generalization. In comparison, the performance gap becomes smaller on the Unseen Map split because all methods are challenged by completely unseen scene layouts. Nevertheless, DynFly still achieves the best overall performance, demonstrating good generalization to unseen environments.

Unlike previous methods that mainly improve navigation through visual representation, spatial modeling, or high-level reasoning, DynFly focuses on the motion generation process between high-level navigation intent and continuous UAV trajectories. As a result, the proposed method improves both object-level generalization on Unseen Object and environment-level generalization on Unseen Map. These results further demonstrate that the proposed dynamic-aware motion interface effectively improves the conversion from high-level navigation intent to executable continuous UAV trajectories and enhances generalization across different unseen scenarios.

\subsection{Ablation Studies}
\subsubsection{Ablation on Different Components}

\begin{table*}[t]
\centering
\caption{Ablation study of key trajectory modeling components across Full/Easy/Hard splits on the Test Unseen Set.}
\label{tab:component_ablation}
\scriptsize
\setlength{\tabcolsep}{2.5pt}
\renewcommand{\arraystretch}{1.10}
\begin{adjustbox}{max width=\textwidth}
\begin{tabular}{lccc cccccc cccccc cccccc}
\toprule
\multirow{2}{*}{Variant}
& \multirow{2}{*}{B-spline}
& \multirow{2}{*}{DiT}
& \multirow{2}{*}{Dyn. Loss}
& \multicolumn{6}{c}{\textbf{Full}}
& \multicolumn{6}{c}{\textbf{Easy}}
& \multicolumn{6}{c}{\textbf{Hard}} \\
\cmidrule(lr){5-10} \cmidrule(lr){11-16} \cmidrule(lr){17-22}
& & &
& NDTW$\uparrow$ & SDTW$\uparrow$ & NE$\downarrow$ & SR$\uparrow$ & OSR$\uparrow$ & SPL$\uparrow$
& NDTW$\uparrow$ & SDTW$\uparrow$ & NE$\downarrow$ & SR$\uparrow$ & OSR$\uparrow$ & SPL$\uparrow$
& NDTW$\uparrow$ & SDTW$\uparrow$ & NE$\downarrow$ & SR$\uparrow$ & OSR$\uparrow$ & SPL$\uparrow$ \\
\midrule
MLP Head
& -- & -- & --
& 12.30 & 6.38 & 92.87 & 23.69 & 43.54 & 20.22
& 11.39 & 5.68 & 68.28 & 22.76 & 42.84 & 18.18
& 13.11 & 7.01 & 114.73 & 24.52 & 44.17 & 22.04 \\
B-spline Head
& \checkmark & -- & --
& 13.72 & 7.60 & 92.43 & 24.51 & 44.30 & 21.21
& 13.04 & 6.91 & 69.40 & 23.96 & 43.51 & 19.68
& 14.32 & 8.05 & 112.91 & 25.00 & 45.00 & 22.58 \\
B-spline DiT
& \checkmark & \checkmark & --
& 13.91 & 8.11 & 89.35 & 27.28 & 45.49 & 22.63
& 12.69 & 7.03 & 66.50 & 25.44 & 45.11 & 19.87
& 14.99 & 9.08 & 109.67 & \textbf{28.93} & 45.83 & 25.09 \\
\textbf{DynFly Full}
& \checkmark & \checkmark & \checkmark
& \textbf{15.89} & \textbf{8.90} & \textbf{83.31} & \textbf{27.60} & \textbf{49.28} & \textbf{23.57}
& \textbf{14.21} & \textbf{7.96} & \textbf{64.85} & \textbf{26.10} & \textbf{46.72} & \textbf{21.26}
& \textbf{17.38} & \textbf{9.73} & \textbf{99.73} & \textbf{28.93} & \textbf{51.55} & \textbf{25.63} \\
\bottomrule
\end{tabular}
\end{adjustbox}
\end{table*}

As shown in Table~\ref{tab:component_ablation}, we progressively evaluate the contribution of each component in DynFly, including the B-spline representation, the Spline-DiT generator, and the dynamic-aware loss. Replacing the conventional MLP trajectory head with a B-spline head consistently improves navigation performance across all three evaluation splits. On the Full split, NDTW increases from 12.30 to 13.72, SDTW improves from 6.38 to 7.60, and both SR and SPL also increase. After introducing Spline-DiT, the performance is further improved. On the Full split, SR increases to 27.28\%, NE is reduced from 92.43\,m to 89.35\,m, and NDTW, SDTW, and SPL also improve. These results show that the B-spline representation provides a better continuous trajectory representation, while the generative trajectory model further learns a more suitable trajectory distribution for UAV navigation.

After adding the dynamic-aware loss, DynFly achieves the best overall performance on most metrics. On the Full split, NDTW increases to 15.89, NE is reduced to 83.31\,m, OSR reaches 49.28\%, and SPL increases to 23.57\%. On the Hard split, DynFly also achieves the best NDTW, SDTW, NE, OSR, and SPL, while SR is tied with the B-spline DiT model. Overall, all three components contribute to the final performance. The B-spline representation models continuous trajectories, Spline-DiT improves trajectory generation, and the dynamic-aware loss further constrains the motion properties of the generated trajectory. Together, they form the proposed dynamic-aware motion interface, which better converts high-level navigation intent into continuous and executable UAV trajectories.

\subsubsection{Sensitivity Analysis on Control-Point}

\begin{figure}[t]
    \centering
    \includegraphics[width=\linewidth]{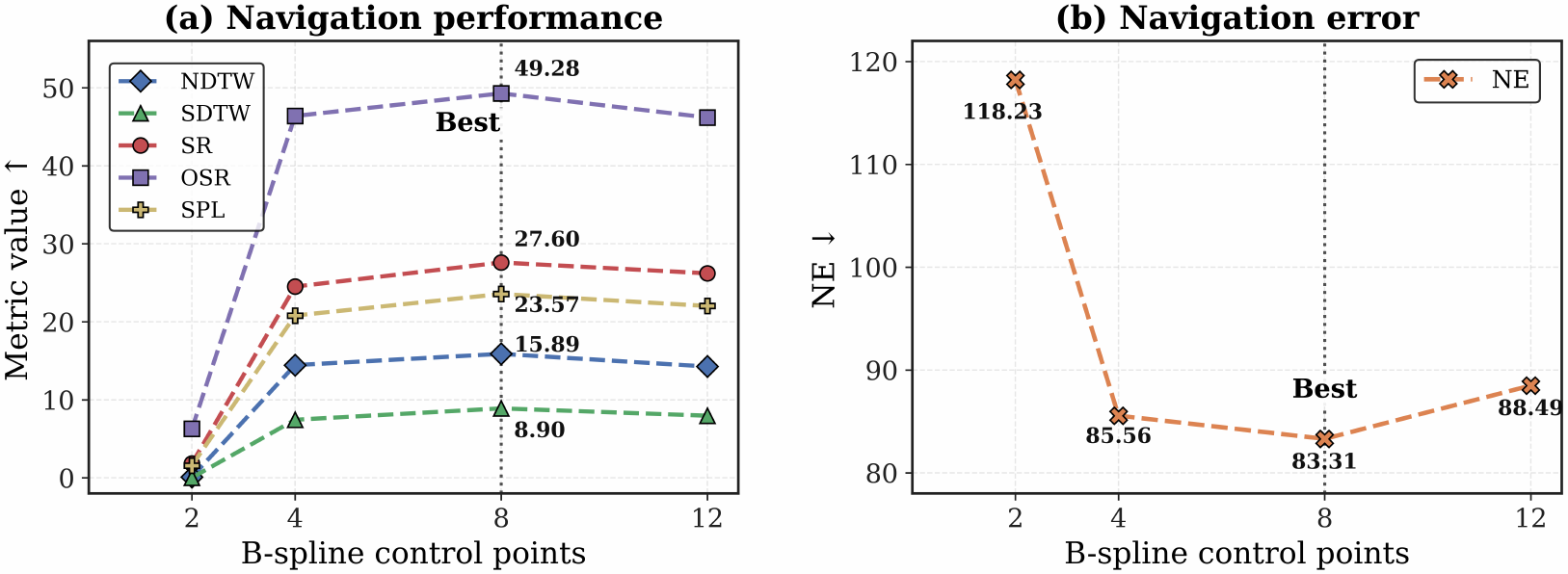}
    \caption{Sensitivity to the number of B-spline control points. Eight control points provide the best overall balance between trajectory capacity and generalization.}
    \label{fig:ctrl_sensitivity}
\end{figure}

Figure~\ref{fig:ctrl_sensitivity} shows the effect of the number of B-spline control points on navigation performance. When only two control points are used, all navigation metrics are much worse and the navigation error is the highest. This indicates that too few control points cannot provide sufficient trajectory representation to describe continuous UAV motion, including local motion changes and turning behaviors. Increasing the number of control points to four leads to clear improvements in all metrics, showing that a richer trajectory representation helps improve navigation performance.

When the number of control points is further increased to eight, the model achieves the best overall performance. NDTW, SDTW, SR, OSR, and SPL all reach their highest values, while NE is reduced to its lowest value. This suggests that eight control points provide a good balance between trajectory representation and model complexity. Increasing the number of control points to twelve causes most metrics to decrease. A possible reason is that more control points increase the flexibility of the trajectory representation and enlarge the control-point search space, making the trajectory distribution more difficult to learn with the current amount of training data. As a result, the overall generalization performance decreases. Therefore, eight control points are used as the default configuration in DynFly.

As shown in Table~\ref{tab:ablation_dynamics_strength}, we study the effect of different dynamic-aware loss weights on navigation performance. Without the dynamic-aware loss (No Dyn), the model is able to generate continuous trajectories, but there is still room for improvement in navigation performance. Using smaller loss weights (Weak Dyn) does not improve the results, and several metrics even decrease. For example, on the Full split, SR drops from 27.28\% to 24.70\%, while NE increases from 89.35\,m to 94.50\,m. This indicates that weak dynamic constraints are not sufficient to effectively guide trajectory generation or fully exploit motion information such as velocity, acceleration, and heading.

\subsubsection{Sensitivity Analysis on Dynamic-Aware Loss}
\begin{table*}[t]
\centering
\caption{Effect of dynamics regularization strength across Full/Easy/Hard splits.}
\label{tab:ablation_dynamics_strength}
\scriptsize
\setlength{\tabcolsep}{2.5pt}
\renewcommand{\arraystretch}{1.10}
\begin{adjustbox}{max width=\textwidth}
\begin{tabular}{l ccc cccccc cccccc cccccc}
\toprule
\multirow{2}{*}{Setting}
& \multirow{2}{*}{$\lambda_v$}
& \multirow{2}{*}{$\lambda_a$}
& \multirow{2}{*}{$\lambda_h$}
& \multicolumn{6}{c}{\textbf{Full}}
& \multicolumn{6}{c}{\textbf{Easy}}
& \multicolumn{6}{c}{\textbf{Hard}} \\
\cmidrule(lr){5-10} \cmidrule(lr){11-16} \cmidrule(lr){17-22}
& & & 
& NDTW$\uparrow$ & SDTW$\uparrow$ & NE$\downarrow$ & SR$\uparrow$ & OSR$\uparrow$ & SPL$\uparrow$
& NDTW$\uparrow$ & SDTW$\uparrow$ & NE$\downarrow$ & SR$\uparrow$ & OSR$\uparrow$ & SPL$\uparrow$
& NDTW$\uparrow$ & SDTW$\uparrow$ & NE$\downarrow$ & SR$\uparrow$ & OSR$\uparrow$ & SPL$\uparrow$ \\
\midrule

No Dyn
& 0 & 0 & 0
&13.91 &8.11 &89.35 &27.28 &45.49 &22.63 
&12.69 &7.03 &66.50 &25.44 &45.11 &19.87 
&14.99 &9.08 &109.67 &28.93 &45.83 &25.09 \\

Weak Dyn
& 0.1 & 0.025 & 0.025

& 13.29 & 7.53 & 94.50 & 24.70 & 42.66 & 20.89
& 12.81 & 7.23 & 65.42 & 25.03 & 45.25 & 19.99
& 13.72 & 7.79 & 120.37 & 24.40 & 40.36 & 21.70 \\

Base Dyn
& 0.2 & 0.05 & 0.05
& \textbf{15.89} & \textbf{8.90} & \textbf{83.31} & \textbf{27.60} & \textbf{49.28} & \textbf{23.57}
& \textbf{14.21} & \textbf{7.96} & 64.85 & \textbf{26.10} & \textbf{46.72} & \textbf{21.26}
& \textbf{17.38} & \textbf{9.73} & \textbf{99.73} & \textbf{28.93} & \textbf{51.55} & \textbf{25.63} \\

Strong Dyn
& 0.4 & 0.1 & 0.1
& 14.08 & 7.42 & 88.89 & 25.27 & 45.56 & 21.16  
& 13.35 & 6.65 & 66.57 & 23.56 & 45.11 & 18.51  
& 14.72 & 8.11 & 108.73 & 26.79 & 45.95 & 23.52  \\

\bottomrule
\end{tabular}
\end{adjustbox}
\end{table*}

The default setting (Base Dyn) achieves the best overall performance. Compared with No Dyn, it improves NDTW from 13.91 to 15.89, SDTW from 8.11 to 8.90, reduces NE from 89.35\,m to 83.31\,m, and increases OSR from 45.49\% to 49.28\% and SPL from 22.63\% to 23.57\% on the Full split. When the loss weights are further increased (Strong Dyn), most navigation metrics decrease again. For example, SR drops to 25.27\% and SPL decreases to 21.16\%. These results suggest that moderate dynamic-aware supervision provides effective motion guidance while preserving the optimization of the navigation objective. In contrast, excessively large loss weights place too much emphasis on motion constraints and reduce navigation performance. Therefore, the Base Dyn setting provides the best balance between motion guidance and navigation performance, and is adopted as the default configuration in DynFly.

To further understand why Base Dyn achieves the best navigation performance, we additionally compare the trajectory quality under different dynamics regularization strengths, as shown in Table~\ref{tab:traj_quality_dyn}. Overall, Base Dyn consistently produces smoother and more stable trajectories across the Full, Easy, and Hard splits. On the Full split, SmoothMean decreases from 0.249 to 0.198, SmoothVar is reduced from 35.359 to 4.759, and TurnMean decreases from 0.185 to 0.158. Similar improvements are also observed on the Hard split, where SmoothVar decreases from 47.541 to 3.179 and TurnMean decreases from 0.183 to 0.156. These results indicate that the dynamic-aware loss effectively suppresses trajectory oscillation and produces smoother continuous flight trajectories.

Compared with No Dyn, Base Dyn consistently reduces both trajectory oscillation and turning variation across all splits. Increasing the loss weights further (Strong Dyn) still maintains relatively stable turning behavior. For example, TurnVar on the Full split decreases slightly from 0.036 to 0.035. However, SmoothVar increases substantially from 4.759 to 26.718, indicating a clear degradation in overall trajectory smoothness. This suggests that overly strong motion constraints reduce the flexibility of trajectory generation, making it more difficult for the model to adapt its trajectory to different navigation scenarios. Overall, moderate dynamic-aware regularization provides a better balance between trajectory quality and navigation performance, which explains why Base Dyn achieves the best overall results.

\begin{table*}[t]
\centering
\caption{Trajectory quality under different dynamics regularization strengths across Full/Easy/Hard splits.}
\label{tab:traj_quality_dyn}
\resizebox{\textwidth}{!}{
\begin{tabular}{lcccccccccccc}
\toprule
\multirow{2}{*}{Setting}
& \multicolumn{4}{c}{Full}
& \multicolumn{4}{c}{Easy}
& \multicolumn{4}{c}{Hard} \\
\cmidrule(lr){2-5} \cmidrule(lr){6-9} \cmidrule(lr){10-13}
& SmoothMean$\downarrow$ & SmoothVar$\downarrow$ & TurnMean$\downarrow$ & TurnVar$\downarrow$
& SmoothMean$\downarrow$ & SmoothVar$\downarrow$ & TurnMean$\downarrow$ & TurnVar$\downarrow$
& SmoothMean$\downarrow$ & SmoothVar$\downarrow$ & TurnMean$\downarrow$ & TurnVar$\downarrow$ \\
\midrule
No Dyn
& 0.249 & 35.359 & 0.185 & 0.044
& 0.255 & 10.825 & 0.190 & 0.051
& 0.247 & 47.541 & 0.183 & 0.041 \\

Base Dyn
& \textbf{0.198} & \textbf{4.759} & \textbf{0.158} & 0.036
& \textbf{0.215} & \textbf{7.984} & \textbf{0.164} & 0.042
& \textbf{0.190} & \textbf{3.179} & \textbf{0.156} & \textbf{0.032} \\

Strong Dyn
& 0.219 & 26.718 & 0.162 & \textbf{0.035}
& 0.235 & 37.983 & 0.166 & \textbf{0.040}
& 0.211 & 21.120 & 0.160 & 0.033 \\
\bottomrule
\end{tabular}
}
\end{table*}

\subsection{Integration with Existing UAV-VLN Frameworks}
\begin{table}[t]
\caption{Compatibility evaluation of DynFly on SpatialFly.}
\label{tab:dyn_module}
\centering
\scriptsize
\setlength{\tabcolsep}{4pt}
\begin{adjustbox}{max width=\linewidth}
\begin{tabular}{lcccccc}
\toprule
Method & NDTW$\uparrow$ & SDTW$\uparrow$ & NE$\downarrow$ & SR$\uparrow$ & OSR$\uparrow$ & SPL$\uparrow$ \\
\midrule
SpatialFly~\cite{jiang2026spatialflygeometryguidedrepresentationalignment}  & 11.20 & 6.50 & 87.82 & 25.46 & 44.41 & 21.76\\
SpatialFly~\cite{jiang2026spatialflygeometryguidedrepresentationalignment} +Dyn & 13.76 & 7.71 & 82.54 & 27.60 & 48.77 & 23.65\\
\midrule
Improvement & +2.56 & +1.21 & -5.28 & +2.14 & +4.36 & +1.89 \\
\bottomrule
\end{tabular}
\end{adjustbox}
\end{table}

To evaluate the compatibility of DynFly with an existing UAV-VLN framework, we integrate it into SpatialFly~\cite{jiang2026spatialflygeometryguidedrepresentationalignment} while keeping the original visual-language front-end and training configuration unchanged. Only the continuous trajectory generation module is replaced. As shown in Table~\ref{tab:dyn_module}, the proposed module consistently improves all navigation metrics. On the Full split, NDTW increases from 11.20 to 13.76, SDTW increases from 6.50 to 7.71, NE is reduced from 87.82\,m to 82.54\,m, and SR, OSR, and SPL improve by 2.14\%, 4.36\%, and 1.89\%, respectively.

These results indicate that the performance improvement mainly comes from the proposed dynamics-aware trajectory generation framework rather than changes to the visual-language backbone. Furthermore, the proposed framework can be integrated into an existing UAV-VLN system while preserving its original visual-language reasoning pipeline. The results demonstrate the effectiveness of explicitly modeling continuous UAV motion between high-level navigation reasoning and flight execution.

\subsection{Generated Trajectory Analysis}

\subsubsection{Qualitative Navigation Comparison}

\begin{figure*}[t]
    \centering
    \includegraphics[width=0.85\linewidth]{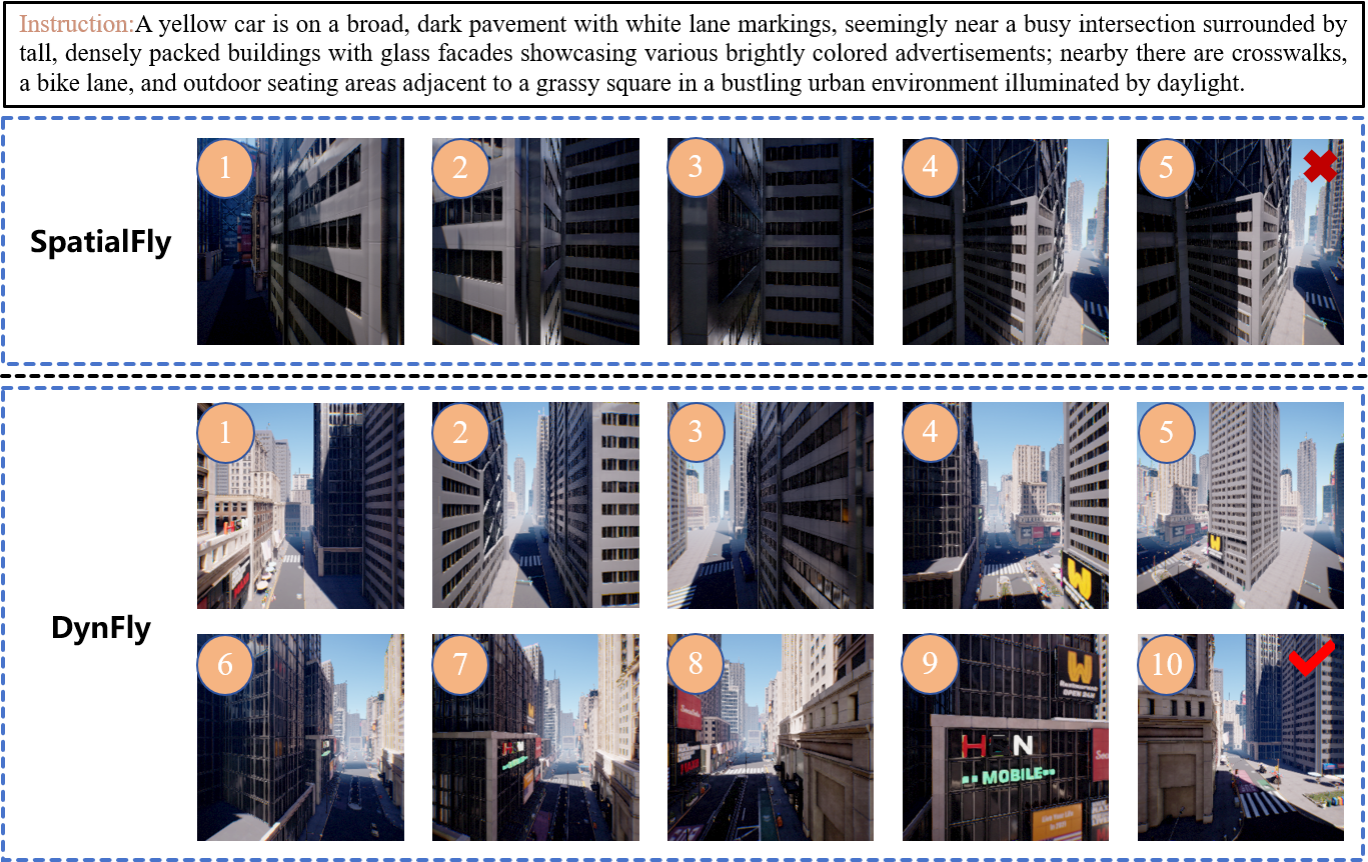}
    \caption{
    Qualitative navigation comparison under the same instruction and visual scene.
    SpatialFly terminates early before reaching the target, while DynFly successfully reaches the target by generating a longer continuous trajectory.
    }
    \label{fig:qualitative_exp}
\end{figure*}

Figure~\ref{fig:qualitative_exp} shows a qualitative navigation example. Under the same navigation instruction and visual observation, SpatialFly terminates early and fails to reach the target region, while DynFly maintains continuous flight and successfully reaches the target. This observation is consistent with the quantitative results and further demonstrates that DynFly improves navigation success in challenging scenes.

From the navigation process, it can be seen that DynFly generates a longer continuous trajectory and maintains a more stable flight direction throughout the navigation process, without obvious early termination or local failure. This suggests that the proposed dynamic-aware motion interface better transforms high-level navigation intent into continuous executable UAV trajectories, leading to more stable navigation behavior.

\subsubsection{Multi-view Trajectory Visualization}

\begin{figure*}[t]
    \centering
    \includegraphics[width=0.8\linewidth]{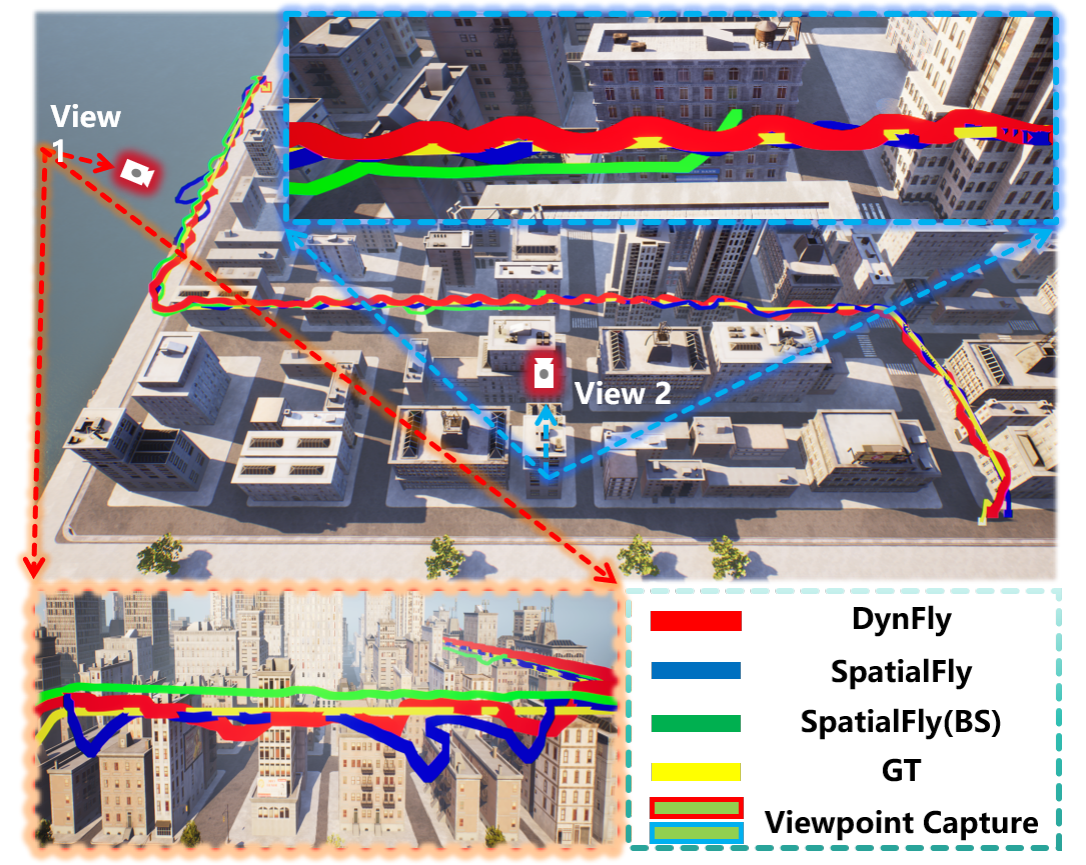}
    \caption{Additional qualitative comparison with multi-view trajectory details. The enlarged views show that DynFly produces a more coherent local flight path under the same visual-language front-end, while the baseline trajectory exhibits larger local oscillations and less stable progress toward the target. This supports that the improvement comes from the proposed dynamic-aware trajectory layer rather than a change of VLM backbone.}
    \label{fig:qualitative_exp2}
\end{figure*}

Figure~\ref{fig:qualitative_exp2} further visualizes the generated trajectories from an overall view and two enlarged local views (View 1 and View 2), together with the ground-truth (GT) trajectory. From the overall view, DynFly generates trajectories that are closer to the GT and maintains continuous flight toward the target. In contrast, both SpatialFly and SpatialFly(BS) show noticeable trajectory deviations and local oscillations in several regions. Overall, DynFly produces more continuous trajectories and follows the target direction more consistently.

View~1 shows the trajectory details at the early stage of flight. In this region, SpatialFly(BS) exhibits obvious local oscillations, indicating that the basic trajectory prediction strategy cannot maintain stable continuous motion. SpatialFly reduces the trajectory deviation by improving spatial representation, but local oscillations are still observed. In comparison, DynFly generates smoother and more continuous trajectories, and its flight direction is more consistent with the GT. This indicates that the proposed dynamic-aware motion interface improves trajectory continuity and reduces unnecessary local motion fluctuations.

View~2 shows the trajectory details when the UAV approaches the target region. In this region, SpatialFly(BS) collides with a building, causing the navigation process to terminate early. SpatialFly avoids the obstacle but still shows a relatively large local trajectory deviation from the GT. In contrast, DynFly successfully avoids the obstacle while maintaining a continuous and stable trajectory that is much closer to the GT. This suggests that DynFly performs more effective local path adjustment in complex environments, improving trajectory executability and navigation stability near the target region.

Overall, the trajectory quality improves progressively from SpatialFly(BS) to SpatialFly and then to DynFly. SpatialFly improves the overall navigation direction through better spatial representation, while DynFly further introduces a dynamic-aware motion interface to model the generation of continuous flight trajectories from high-level navigation intent. As a result, the model not only predicts where the UAV should fly, but also learns how to generate continuous trajectories that better match UAV flight characteristics. Under the same visual-language front-end, DynFly further reduces the gap between the predicted trajectory and the ground-truth trajectory. These observations are consistent with the quantitative results, the trajectory-quality analysis, and the ablation studies, further demonstrating the effectiveness of DynFly for dynamic-aware continuous trajectory generation.

\section{Conclusion}
\label{sec:conclusion}

This paper presented DynFly, a dynamics-aware continuous trajectory generation framework for UAV vision-language navigation. Instead of improving the visual-language front-end, DynFly focuses on bridging high-level navigation reasoning and executable UAV motion. By introducing a B-spline-based trajectory representation, a Spline-DiT trajectory generator, and dynamic-aware supervision, DynFly generates continuous UAV trajectories that better match UAV motion characteristics. Experimental results show that the proposed framework improves both navigation performance and trajectory quality, while the ablation studies further verify the effectiveness of the B-spline trajectory representation, the Spline-DiT generator, and the dynamic-aware supervision. More importantly, the results demonstrate that explicitly modeling continuous UAV motion not only improves trajectory quality, but also serves as an effective bridge between high-level navigation reasoning and low-level flight execution, leading to better navigation performance in complex and unseen environments.

\textbf{Limitations and future work.} Although DynFly improves both navigation performance and trajectory quality, this work still has some limitations. First, the proposed framework is evaluated only in simulation environments, and its performance on real UAV platforms still needs to be verified. Second, the current framework focuses on short-horizon trajectory generation from high-level navigation intent. The connection with low-level flight control and long-horizon navigation has not been studied yet. In addition, further improvements in model lightweighting, real-time inference, and sim-to-real transfer are needed for practical applications. In the future, we will evaluate DynFly on real UAV platforms, improve its efficiency, and explore its application to more UAV vision-language navigation and vision-language-action systems.


\bibliographystyle{IEEEtran}

\bibliography{references}






\end{document}